\documentclass[letterpaper]{article} % DO NOT CHANGE THIS
\usepackage{aaai24}  % DO NOT CHANGE THIS
\usepackage{times}  % DO NOT CHANGE THIS
\usepackage{helvet}  % DO NOT CHANGE THIS
\usepackage{courier}  % DO NOT CHANGE THIS
\usepackage[hyphens]{url}  % DO NOT CHANGE THIS
\usepackage{graphicx} % DO NOT CHANGE THIS
\usepackage{natbib}  % DO NOT CHANGE THIS AND DO NOT ADD ANY OPTIONS TO IT
\usepackage{caption} % DO NOT CHANGE THIS AND DO NOT ADD ANY OPTIONS TO IT
\usepackage{subcaption}
\usepackage{algorithm}
\usepackage{algorithmic}

\usepackage{amsmath}
\usepackage{multirow}
\usepackage{booktabs}
\usepackage{bbding}
\usepackage{xspace}
\usepackage{amssymb}
\providecommand*{\onedot}{\futurelet\@let@token\@onedot}
\providecommand*{\eg}{e.g.,~}

\providecommand*{\ie}{i.e.,~}

\providecommand*{\etal}{et~al\onedot}

\usepackage{newfloat}
\usepackage{listings}
\DeclareCaptionStyle{ruled}{labelfont=normalfont,labelsep=colon,strut=off} % DO NOT CHANGE THIS
\floatstyle{ruled}
\newfloat{listing}{tb}{lst}{}
\floatname{listing}{Listing}
\title{NeRF-LiDAR: Generating Realistic LiDAR Point Clouds with Neural Radiance Fields}
\author {
    Junge Zhang\textsuperscript{\rm 1}, \
    Feihu Zhang\textsuperscript{\rm 2}, \
    Shaochen Kuang\textsuperscript{\rm 3}, \
    Li Zhang \textsuperscript{\rm 1}\thanks{Li Zhang (lizhangfd@fudan.edu.cn) is the corresponding author with School of Data Science, Fudan University.} \\
}
\affiliations {
    \textsuperscript{\rm 1}Fudan University \quad
    \textsuperscript{\rm 2}University of Oxford \quad
    \textsuperscript{\rm 3}South China University of Technology \\
\vspace{2mm} 
    \url{https://github.com/fudan-zvg/NeRF-LiDAR}
}

\begin{document}

\maketitle

%%%%%%%%% ABSTRACT
\begin{abstract}
Labelling LiDAR point clouds for training autonomous driving is extremely expensive and difficult. LiDAR simulation aims at generating realistic LiDAR data with labels for training and verifying self-driving algorithms more efficiently. 
Recently, Neural Radiance Fields (NeRF) have
been proposed for novel view synthesis  using implicit reconstruction of 3D scenes. 
Inspired by this, we present NeRF-LIDAR, a novel LiDAR simulation method that leverages real-world information to generate realistic LIDAR point clouds. Different from existing LiDAR simulators, we use real images and point cloud data collected by self-driving cars to learn the 3D scene representation, point cloud generation and label rendering. 
We verify the effectiveness of our NeRF-LiDAR  by training different 3D segmentation models on the generated LiDAR point clouds. 
It reveals that the trained models are able to achieve similar accuracy when compared with the same model trained on the real LiDAR data.  Besides, the generated data is capable of  boosting the accuracy through pre-training which helps reduce the requirements of the real labeled data. 
Code is available at \url{https://github.com/fudan-zvg/NeRF-LiDAR}

\end{abstract}

%%%%%%%%% BODY TEXT
\section{Introduction}
LiDAR sensor plays a crucial role in autonomous driving cars for 3D perception  and planning. However, labelling the 3D point clouds for training 3D perception models is extremely expensive and difficult.  
In view of this, LiDAR simulation that aims at generating realistic LiDAR point clouds for different types of LiDAR sensors becomes increasingly important for autonomous driving cars.  It can generate useful LiDAR data with labels for developing and verifying the self-driving system.

\begin{figure}[ht!]
\centering
\vspace{-4mm}
\begin{subfigure}[t]{0.485\linewidth}

%\begin{minipage}[t]{0.5\linewidth}
%\centering
\includegraphics[width=1\linewidth,angle=270]{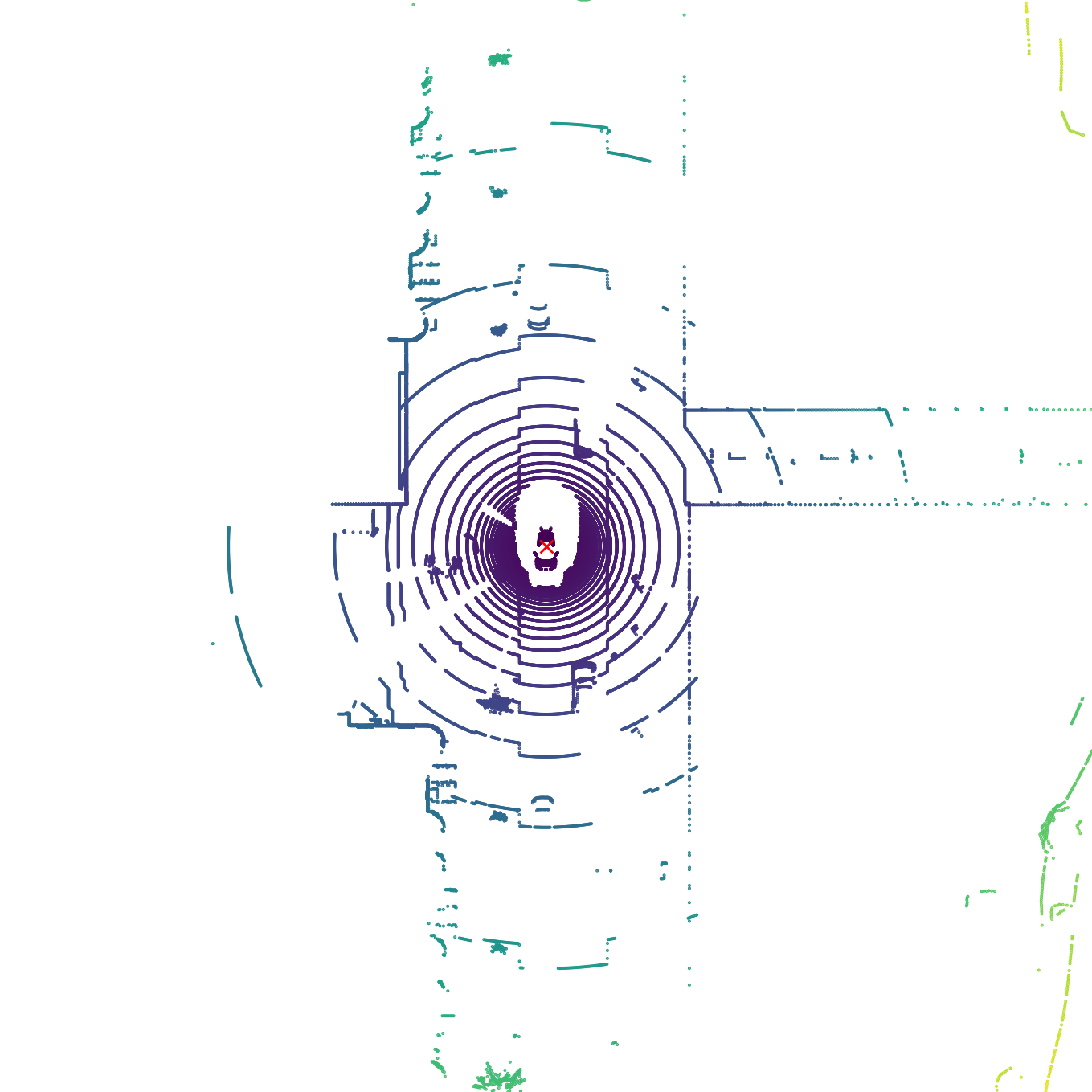}
\caption{CARLA}
\label{subfig:carla}
\end{subfigure}
%\end{minipage}%
\begin{subfigure}[t]{0.485\linewidth}
\vspace{0pt}
\centering
\includegraphics[width=1\linewidth,angle=270]{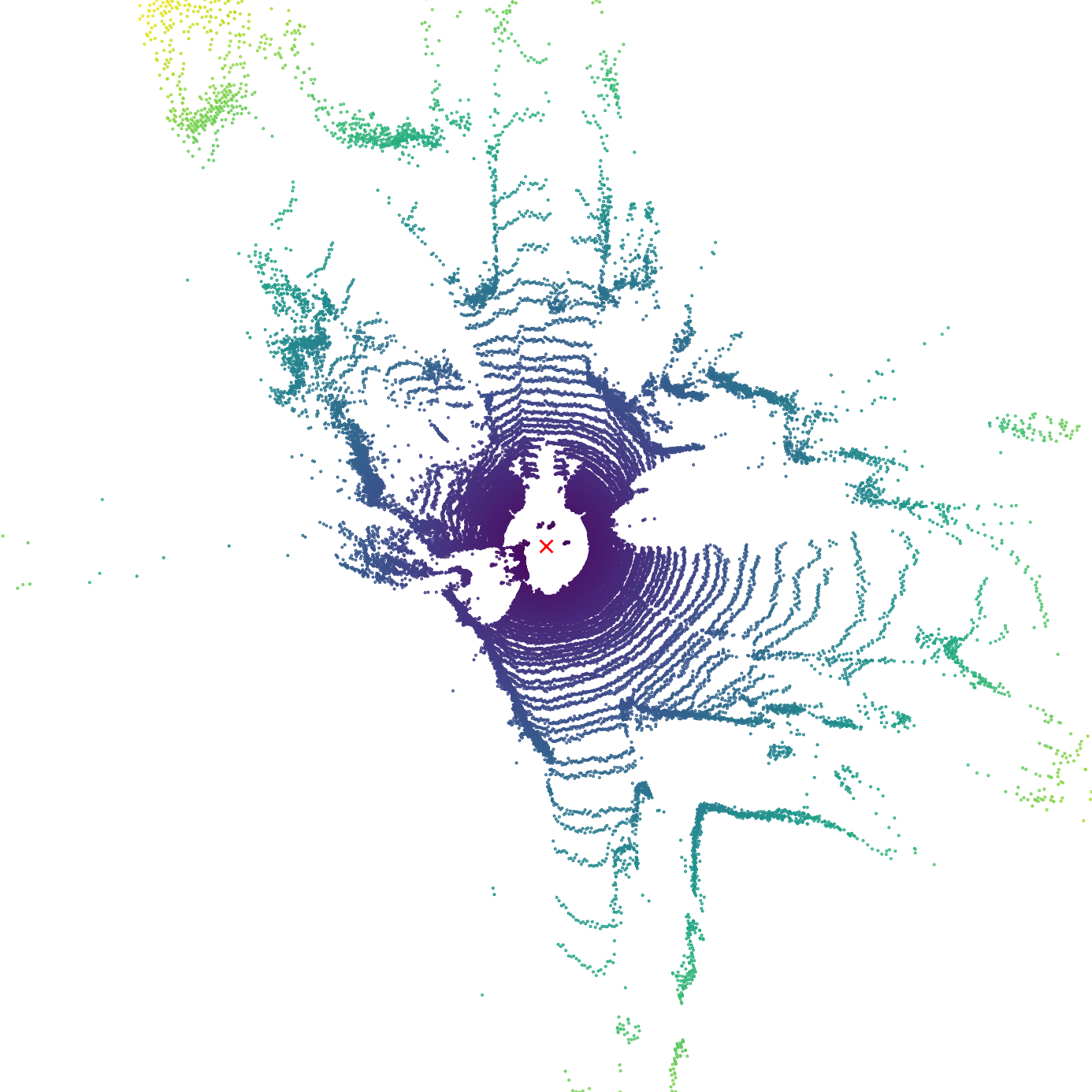}
%\caption{fig1}
\caption{LiDARGen}
\end{subfigure}\\
\begin{subfigure}[t]{0.485\linewidth}
\centering
\includegraphics[width=1\linewidth]{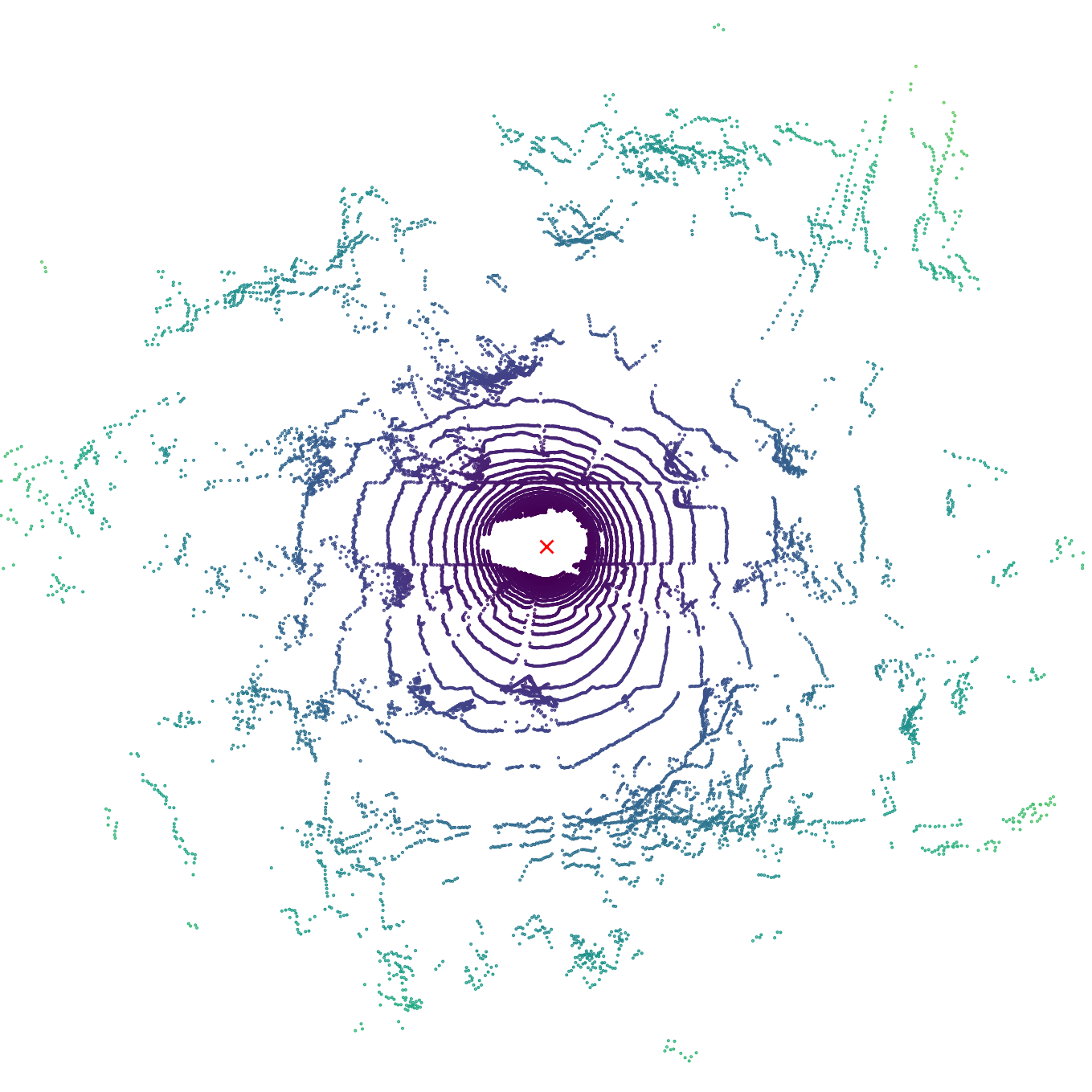}
\caption{Our NeRF-LiDAR}
\label{subfig:our_nerf_lidar}
\end{subfigure}%
%\hspace{-2mm}
\begin{subfigure}[t]{0.485\linewidth}
\centering
\includegraphics[width=1\linewidth]{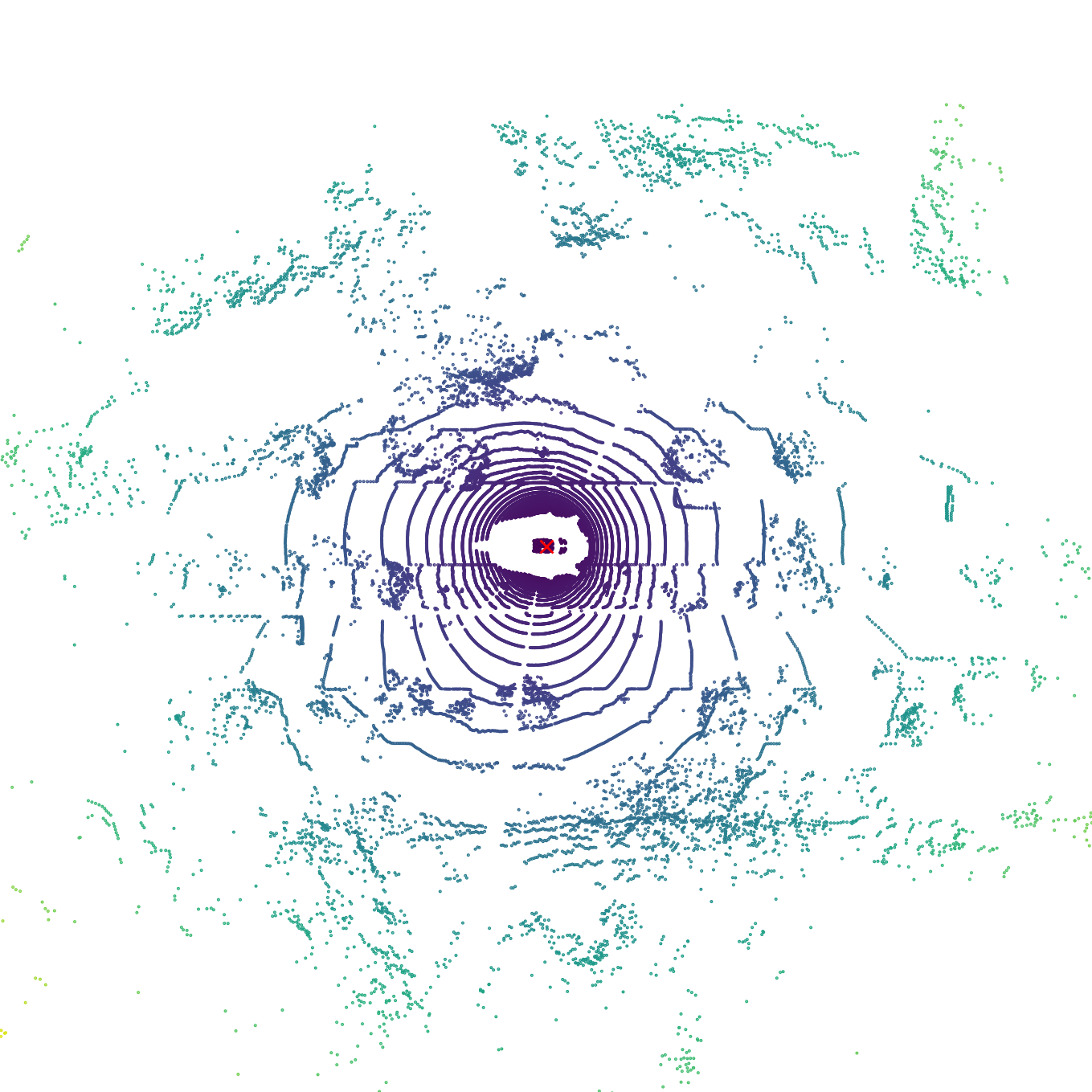}
\caption{Real LiDAR sensor}
\end{subfigure}%

\caption{Comparisons of results between our NeRF-LiDAR and other existing LiDAR simulation methods. (a) Method \cite{carla} that creates virtual world for LiDAR simulation. (b) Diffusion model used for LiDAR generation \cite{lidargen}. (c) Our NeRF-LiDAR can generate realistic point clouds that is nearly the same as the real LiDAR point clouds (d).}
\label{fig:performance_illustration}
\vspace{-6mm}
\end{figure}

Many previous works have studied the LiDAR simulation, which can be mainly categorized into two types: the 
virtual environment creation method  and the reconstruction-based method. The former creates the 3D virtual world by graphics-based 3D modeling and then generates the 3D LiDAR point clouds by physics-based simulation (ray tracing). These kinds of works \cite{carla,gazebo} have natural limitations as it's impossible for 3D modelers to create a virtual world that is the same as the complex real world. The  simulated LiDAR points have significant domain differences from the real LiDAR points and cannot be used to train robust deep neural network models.
The latter \cite{lidarsim,fang2020augmented} relies on multiple LiDAR scans to densely reconstruct the street background and then place the foreground objects into the background. However, it's expensive to collect dense LiDAR scans which may need special devices \cite{fang2020augmented}. Moreover, it's expensive to generate point-wise semantic labels for simulated LiDAR data. It still requires human annotations on the 3D scans.

Recently, Neural Radiance Fields (NeRF) \cite{NeRF,Mip-NeRF} have been proposed for implicit reconstruction of the 3D object/scenes  with multiple images as inputs. NeRF can render photo-realistic novel views along with dense depth maps. 

Inspired by this, we proposed to learn a NeRF representation for real-world scenes and render LiDAR point clouds along with accurate semantic labels. Different from existing reconstruction-based LiDAR-simulation methods \cite{lidarsim,fang2020augmented} or the virtual world creation \cite{carla} (Fig. \ref{subfig:carla}), 
our method takes full use of the multi-view images to implicitly reconstruct the labels and 3D real-world spaces. The multi-view images can assist the simulation system to learn more accurate 3D geometry and  real-world details and generate more accurate point labels.  
The proposed NeRF-LiDAR model consists of two important modules: 1) the reconstruction module that uses NeRF to reconstruct the real world along with labels; 2) the generation module that learns to generate realistic point clouds through a point-wise alignment and a feature-level alignment.

Since our NeRF-LiDAR can generate realistic LiDAR point clouds along with accurate semantic labels, 
in the experiments, we verify the effectiveness of our NeRF-LiDAR  by training different 3D segmentation models on the generated LiDAR point clouds. 
The trained 3D segmentation models are shown able to achieve competitive performance when compared with the same model trained on the real LiDAR data which implies that the generated data can be directly used to replace the real labeled LiDAR data. Besides, by using the generated LiDAR data for pre-training and a small number of real data (\eg 1/10) for fine-tuning, the accuracy can be significantly improved by a large margin which is even better than the model trained on a 10 times larger real LiDAR dataset.

\section{Related Work}
LiDAR point-cloud simulation has been studied for many years from the initial engine based rendering to the state-of-the-art real-world reconstruction-based LiDAR rendering.

\paragraph{LiDAR Simulation}

The first type of the LiDAR simulation method \cite{carla,gazebo,v2w_simu,gschwandtner2011blensor} relies on creating 3D virtual world and rendering the point clouds with physics-based simulation. However, the generated virtual data have large domain gaps with the real data when used for training deep neural networks. This is because the 3d virtual
world cannot simulate the complexity and details of the real world.
Another point-cloud simulation method ~\cite{generative_pointcloud,diffusion_pointcloud,pointflow,lidargen}  generates 3D point clouds based on generative models. However, these generated data cannot be used to train deep neural network models as they also have significant domain differences with real point clouds. Moreover, it's difficult to generate labels for the point clouds.

State-of-the-art LiDAR simulation methods \cite{fang2020augmented,lidarsim} first reconstruct the real-world driving scenes into 3D meshes and then run the physics-based simulation. In order to achieve dense accurate reconstruction results, these methods need to scan the street many times using expensive LiDAR devices \cite{fang2020augmented}.  
More importantly, it’s still expensive to generate point-wise semantic
labels for simulated LiDAR data as it requires human
annotations on the reconstructed 3D scenes. 
Instead of simulating the whole LiDAR scenes, others, \eg ~\cite{LiDAR-Aug}  use the real-world 3D scenes and propose a rendering-based LiDAR augmentation framework to enrich the training data and boost the performance of LiDAR-based 3D object detection.
Our method also leverages real-world information for learning LiDAR simulation. Our NeRF-LiDAR creates an implicit neural-radiance-field  representation of the real world for both point clouds and label rendering.

\paragraph{Neural Radiance Fields}
Recently, Neural radiance fields (NeRF) ~\cite{NeRF} have been proposed  as an implicit neural representation of the 3D real world for novel view synthesis. NeRFs can take multiple 2D images and their camera-view directions to represent the whole 3D space. However, early NeRFs are only applicable to small object-centric scenes. 

Many recent NeRFs have been proposed to address the challenges of large-scale outdoor scenes
~\cite{Mip-NeRF,Mip-NeRF360,blocknerf,nerf++}. There are also some methods ~\cite{urban-nerf,dsnerf} leveraging depth supervision to help create more accurate 3D geometry of scenes. 
Panoptic or semantic label synthesis for novel views is also explored in \cite{semanticnerf,panoptic-neural,panoptic-nerf}. They utilize the density of the volume to render image labels along with the novel view synthesis. 
Inspired by these works, our method reconstructs the accurate 3D geometry using the NeRF methodology in the driving scene and generates 3D point clouds along with accurate semantic labels for the LiDAR simulation. 

%------------------------------------------------------------------------

\section{NeRF-LiDAR}
\begin{figure*}[ht]
    \centering
    \setlength{\abovecaptionskip}{0pt}
\setlength{\belowcaptionskip}{-10pt}
\includegraphics[width=1\linewidth,trim={1.5cm 8cm 1.5cm 8cm},clip]{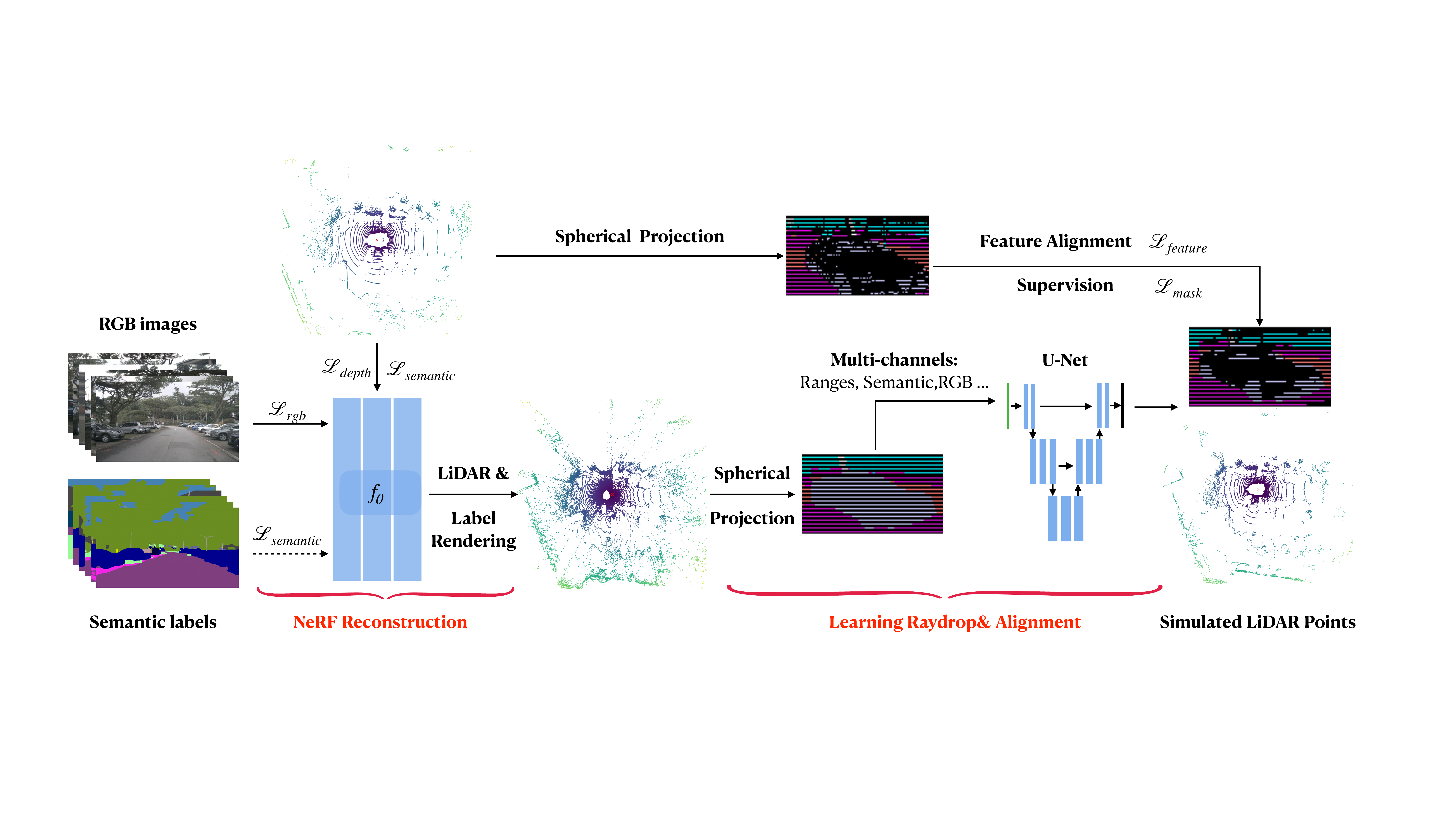}
    \caption{Schematic illustration of NeRF-LiDAR. Image sequences along with the predicted weak semantic labels are used as inputs to reconstruct the implicit NeRF model.  LiDAR signals are also used to help create more accurate 3D geometry. Initial coarse point clouds are generated by the NeRF reconstruction through Eq. \eqref{eq:lidar_direction}$\sim$\eqref{eq:point_generation}. The initial point clouds are projected into 2D equirectangular images.  We then utilize a U-Net to learn raydrop and the alignment (detailed in Fig. \ref{fig:alignment}) to make the generated point clouds more realistic.}
    \label{fig:overview}
    % \vspace{-3mm}
\end{figure*}
In this section, we present our NeRF-based LiDAR simulation framework (as shown in Fig. \ref{fig:overview}). The method consists of three key components: 1) NeRF reconstruction of the driving scenes, 2) realistic LiDAR point clouds generation and 3) point-wise semantic label generation. We formulate the three components into end-to-end deep neural network models for learning LiDAR simulation. 

\paragraph{Neural Radiance Fields}
Neural Radiance Fields learn implicit representation for the scenes and render novel view synthesis through volume rendering. It learns a function $f:(\mathbf{x},\mathbf{v}) \rightarrow (\mathbf{c},\sigma)$ for mapping coordinates $\mathbf{x}$ and viewing directions $\mathbf{v}$ to color $\mathbf{c}$ and density $\sigma$. The volume rendering is based on discrete rays $\mathbf{r} = \mathbf{o}+t \mathbf{d}$ in the space and applying numerical integration along the rays to query color:
\begin{align}
\setlength{\abovecaptionskip}{4pt}
\setlength{\belowcaptionskip}{4pt}
    \hat{\mathbf C}(\textrm{r}) = \sum_{i=1}^NT_i(1-e^{-\sigma_i\delta_i}) \mathbf c_i, \
    T_i = \exp\left( -\sum_{j=1}^{i-1}\sigma_j\delta_j \right) \label{eq:volume_rendering},
\end{align}
where $\mathbf{o}$ is the origin of the ray, $\mathbf{d}$ is the direction, $T_i$ is the accumulated transmittance along the ray, and $\mathbf c_i$ and $\sigma_i$ are the corresponding color and density at the sampled point $t_i$. $\delta_j = t_{j+1}-t_j$ refers to the distance between the adjacent sampled points.

\paragraph{NeRF Reconstruction}
State-of-the-art LiDAR simulation methods \cite{lidarsim} rely on dense LiDAR scans for scene reconstruction. To achieve the dense reconstruction of the street, \cite{LiDAR-Aug} uses a special (expensive) LiDAR device to collect dense depth maps. \cite{lidarsim} scans the street many times to accumulate much denser point clouds. These dense depth maps or point clouds are then used to extract the meshes of the street. Finally, the meshes are used to generate point clouds of different types of LiDAR sensors.

In this paper, we present a new method takes multi-view images and sparse LiDAR signals to reconstruct the street scenes and represent the 3D scenes as an implicit NeRF model. We propose to use the driving-scene data to learn the NeRF reconstruction.

NeRF reconstruction of the unbounded large-scale driving scenes is challenging. This is because most of NeRFs \cite{NeRF} are designed for small scene reconstruction with object-centric camera views. However, the driving data are often collected in the unbounded outdoor scenes without object-centric camera views settings (\eg nuScenes~\cite{nuscenes2019}).
Moreover, since the ego car moves fast during the data collection, the overlaps between adjacent camera views are too small to be effective for building multi-view geometry. 
We reconstruct the NeRF representation based on the multi-view images and leverage the LiDAR points to provide extra depth supervision to create more accurate 3D geometries. Besides, the real LiDAR point clouds are used as supervision to learn more realistic simulated LiDAR data.

To reconstruct the driving scenes, we use the unbounded NeRF \cite{Mip-NeRF360} with a modified supervision of:
\begin{align}
    &\mathcal{L}_{rgb} = \|\hat{\mathbf{C}}(\mathbf{r}) - \mathbf{C}(\mathbf{r})\|_2 ,\\
    &\mathcal{L}_{depth} = \|\hat{\mathcal{D}}(\mathbf{r}) - \mathcal{D}(\mathbf{r})\|_1.
%    &\hat{\mathcal{D}} = \sum_{i=1}^{N}T_i\big(1-\exp(-\sigma_{i}\delta_{i})\big) z_i.  
\end{align}
Here, $\hat{D}$ is the rendered depth by the volume rendering in Eq. \eqref{eq:volume_rendering}:
\begin{equation} \label{eq_render_depth}
    \hat{\mathcal{D}}(\mathbf{r}) = \sum_{i=1}^{N}T_i\big(1- e^{-\sigma_i\delta_i}\big) z_i,
\end{equation} 
where $z_i$ is the depth value  at the sampled point $t_i$ on the ray $\mathbf{r}$. Since the original unbounded NeRF~\cite{Mip-NeRF360} is extremely slow which takes about one day to train each NeRF block, we adopt hash-encoding NeRF~\cite{zipnerf} to speed up the simulation process.

\paragraph{Point-cloud Generation}
After learning the implicit NeRF representation of the driving scenes, we set a virtual LIDAR to simulate the real LIDAR sensor. The virtual LIDAR shares the same parameter settings with the real LiDAR sensors. For example,  nuScenes~\cite{nuscenes2019} uses Velodyne HDL32E LiDAR sensor, of which the spinning speed is 20Hz and the field of view ranges from -30.67 degree to 10.67 degree (consists of 32 channels). 
We can therefore simulate NeRF-LiDAR rays with the LiDAR center ($\mathbf{o}$) and direction $\mathbf{d}$ accordingly:
\begin{align}
    \mathbf{d} &= (\cos\theta \cos \phi, \ \sin \theta \sin \phi, \ \cos \phi)^T,
    \label{eq:lidar_direction}
\end{align}
where $\theta,\phi$ represent the azimuth and vertical angle of the ray, determined by the time interval of the lasers and the settings of the LiDAR sensor. 

The origin of the rays $\mathbf{o}$ changes according to the defined motion of ego cars.
\begin{align}
    \mathbf{o} &= \mathbf{o_0}+\Delta t\cdot \mathbf{v}.
    \label{eq:lidar_orgin}
\end{align}

Here, $\mathbf{v}$ is the velocity of the ego vehicle and $\Delta t$ means the time interval from the previous state.  $\Delta t$ is decided by the frame rate of the LiDAR sensors (\eg 20Hz for nuScenes LiDAR). 

Each simulated point $\mathbf{p}=\{x,y,z\}$ can be then calculated by the pre-defined directions $\mathbf{d}$ of rays and the distances $\hat{D}$ from the LiDAR sensor to the real world objects:

\begin{align}
    \mathbf{p} = \mathbf{o} + \hat{D} \mathbf{d}.
    \label{eq:point_generation}
\end{align}
There are about 20$\sim$40k points in one frame of a standard 32-channel LiDAR point clouds. To simulate the whole point clouds, for each point, we render a ray to compute the exact 3D location.

\paragraph{Label Generation} 
To achieve the point-wise semantic labels of the simulated LiDAR points, we use the 2D semantic labels of the images to learn the 3D label generation. 

 Semantic NeRF \cite{semanticnerf} proposes to use the  semantic logits that could be rendered through volume rendering (Eq. \eqref{eq:volume_rendering}) like RGB color:
\begin{align}
\setlength{\abovecaptionskip}{0pt}
\setlength{\belowcaptionskip}{4pt}
    \hat{\mathbf S}(\textrm{r}) = \sum_{i=1}^NT_i(1-e^{-\sigma_i\delta_i}) \mathbf s_i,
    % , \quad T_i = \exp\left( -\sum_{j=1}^{i-1}\sigma_j\delta_j \right)
    \label{eq:logits_rendering}
\end{align}
where $T_i,\sigma_i,\delta_i$ follows the definition of Eq. \ref{eq:volume_rendering}, $s_i$ is the semantic logit of the sampled point.

Here, we first consider the most difficult cases where there is no annotated label from the collected driving data (images and LiDAR points). Given unlabeled real images collected from multiple cameras of the self-driving cars, 
we train a SegFormer~\cite{segformer} model, on the mixture of other datasets including Cityscapes \cite{cordts2016cityscapes}, Mapillary~\cite{neuhold2017mapillary}, BDD~\cite{bdd100k}, IDD~\cite{varma2019idd} to compute weak labels that serve as inputs to the NeRF reconstruction model.  To achieve better cross-dataset generalization  of the SegFormer and avoid conflicts in the label definition, we utilize the learning settings and label merging strategy in the \cite{mseg}. 

Considering that the generated weak labels may have many outliers, to reduce the influence of these outliers and generate more accurate 3D point labels, we take full use of  multi-view geometric and video spacial temporal consistency in our NeRF reconstruction.

In the NeRF training, we combine the image-label supervision into the reconstruction learning by constructing the semantic radiance fields:
\begin{align}
    \hat{\mathcal{L}_{l}} = CE(\hat{\mathbf{S}}(\mathbf{r}) , \mathbf{S}(\mathbf{r})),
\end{align}
where $CE$ is the cross-entrophy loss, $\mathbf{r}$ represents pixel-wise camera rays corresponding to each image pixel, $\hat{\mathbf{S}}(\mathbf{r})$ is the rendered labels by the NeRF model (Eq.\eqref{eq:logits_rendering}) and $\mathbf{S}(\mathbf{r})$ is the label predicted by the image segmentation model.

In some other cases, when there is a small number of labeled images or LiDAR frames, we can also leverage the existing ground-truth labels for more robust label generation. For example, in the nuScenes dataset, a small part of the LiDAR frames (about 1/10) was labeled with semantic annotations.   We take the sparse 3D point labels along with the weak 2D image labels to learn more accurate semantic radiance fields.
\begin{align}
    \mathcal{L}_{l} = CE(\hat{\mathbf{S}}(\mathbf{r}_{LiDAR}) , \mathbf{S}(\mathbf{r}_{LiDAR})).
\end{align}
Here $\mathbf{r}_{LiDAR}$ represents the point-wise rays emitted by the LIDAR sensor.  
The total loss for learning our NeRF reconstruction can be represented as:
\begin{align}
    \mathcal{L}_{rec} = \mathcal{L}_{depth} \ + \  w_{rgb} \mathcal{L}_{rgb} \ + \ \hat{w_{l}}\hat{\mathcal{L}_{l}} \ + \ \mathcal{L}_{l},
    \label{eq:rec_loss}
\end{align}
where $w_{rgb}$ and $w_{l}$  balance the RGB geometry reconstruction, the LiDAR rendering and the semantic label rendering. 

\paragraph{Learning  Raydrop \& Alignment}
In the real world, the LiDAR sensor cannot receive all beams emitted by itself, influenced by the 
reflectance ratio of different materials (\eg glasses), the incidence angle and many other factors~\cite{lidarsim,fang2020augmented}. Points are usually dropped when the reflected intensity is below the perception threshold. To make generated LIDAR points closer to the real LIDAR points, we learn a raydrop processing on the generated dense points.

NeRF-LiDAR allows us to render depths (3D points) at arbitrary positions and directions.  We use the ground-truth LiDAR frames as supervision to learn the raydrop. Given one ground-truth LiDAR frame $P$, we render the simulated LiDAR frame $\hat{P}$ at the same location accordingly.  The ground-truth $P$ and the simulated $\hat{P}$  should have strong point-wise correspondence. 
\begin{align}
    P \simeq \hat{P}.
\end{align}
We adopt such point-to-point correspondence as the learning target. 

\paragraph{Equirectangular Image Projection}
It's difficult to create a point-to-point correspondence between the two irregular 3D point clouds. To better leverage the point-wise correspondence, we first render all generated points into a 2D equirectangular image (a panorama sparse depth image, as illustrated in Fig. \ref{fig:overview}). For example, in nuScenes dataset, the resolution of the 32-channel LiDAR equirectangular image is set as $32\times 1024$.

\begin{figure}[t]
    \centering
    \includegraphics[width = 1.1\linewidth,trim={14cm 12cm 13cm 6cm}, clip]{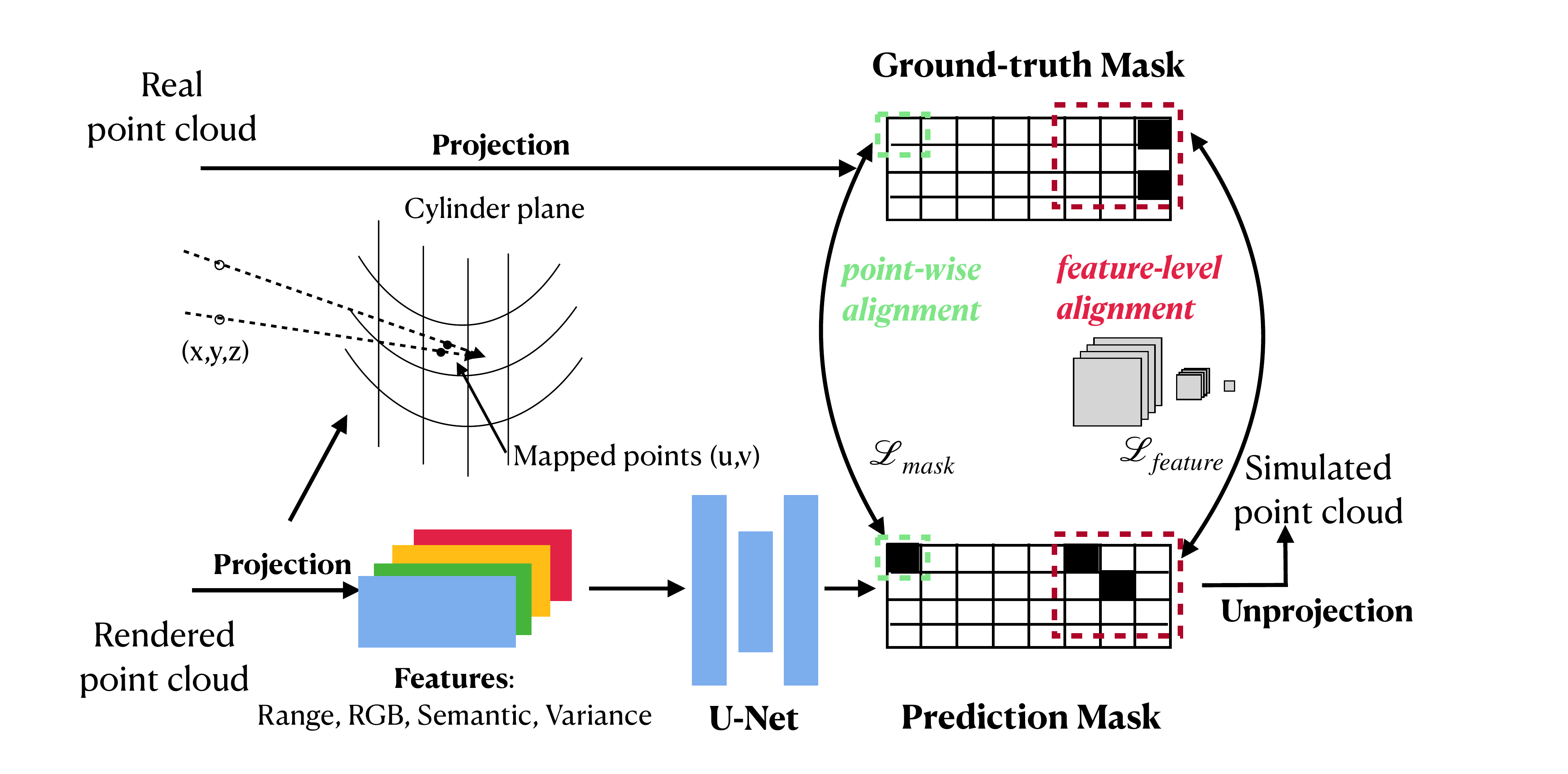}
    \caption{Illustration of learning raydrop and alignment. The initial coarse point clouds are projected into 2D equirectangular images. We use the projected depth, RGB texture, and depth variances as input to a standard U-Net. The U-Net learns the raydrop mask to improve the initial coarse point clouds through the point-wise alignment (Eq. \eqref{eq:point_alignment}) and the feature-level alignment (Eq. \eqref{eq:feature_alignment}). Finally, the refined equirectangular images are back-projected to 3D space to achieve the expected LiDAR point clouds.}
    \label{fig:alignment}
\vspace{-5mm}
\end{figure}

To project the irregular LiDAR points, we adopt the spherical projection ~\cite{semantickitti} to project our points into the equirectangular image grids (as illustrated in Figure \ref{fig:alignment}). 
Similarly, the real LiDAR frame is also transferred into a 2D $32\times 1024$ equirectangular image. In this way, we can easily create the correspondence in 2D grids. 

\paragraph{Point-to-point Alignment}
To learn the drop probabilities for each 2D grid location in the equirectangular image, we employ a standard U-Net which encodes the depth ranges, semantic labels, RGB textures, and depth variances  between neighborhoods into a feature representation. The U-Net outputs a 2D probability map (a binary mask) to represent the raydrop results.  We take the corresponding real LiDAR equirectangular image as the learning target.

\begin{align}
    \mathcal{L}_{mask} = CE(\hat{M},M_{gt}),
    \label{eq:point_alignment}
\end{align}
where $\hat{M},M_{gt}$ represent predicted and true mask respectively.
Extra points/grids are dropped through the learned drop mask. The expected 3D point clouds can be achieved through back-projection of the equirectangular image. 

\paragraph{Feature-level Alignment}
The above point-wise alignment aims at making the generated points more realistic or spatially closer to the real point locations. 
However, the generated LiDAR data will finally be used to train deep neural network models for 3D perceptions. 
To make the generated data more effective and able to achieve better accuracy when training deep neural networks (\eg 3D perception models), we propose the feature-level alignment to further regulate the distribution of the generated point clouds.

\begin{align}
    \mathcal{L}_{feat} = \sum_{k=1}^n w_k \|F(\hat{I})_k - F(I_{gt})_k\|_1, 
    %\mathcal{W}(F(\hat{I}), F(I_{gt}))
    \label{eq:feature_alignment}
\end{align}
where $F$ is a pre-trained and fixed feature extractor (\eg VGG \cite{vgg}, Point-cloud segmentation Net \cite{milioto2019rangenet++}). 
We use $n$-level pyramidal features to compute the feature distance.
$n=4$ is the number of feature levels, $w_k = 2^{k-n}$ is the weights for $k$th-level features. The loss $L_{feat}$ measures the feature-level similarity between the real and the simulated point clouds. To enable the back-propagation from the feature loss to the previous raydrop module, we apply the gumble-softmax \cite{jang2016categorical} on the ray drop processing. 

The whole generation target can be represented as:
\begin{align}
    \mathcal{L}_{gen} =  \mathcal{L}_{mask} +  {w_{feat}} \mathcal{L}_{feat}.
    \label{eq:gen_loss}
\end{align}

The feature-level distribution alignment can make the generated data more effective and achieve better accuracy in training deep segmentation networks. Besides, we find that it also helps to remove extra outliers and floats in the generated point clouds (examples are shown in Fig. \ref{fig:ablations} and the supplementary materials).

\section{Experiment}
\subsection{Experimental Settings}
\paragraph{Dataset}
We use the standard nuScenes self-driving dataset for training and evaluation.  NuScenes contains about 1000 scenes collected from different cities. Each scene consists of about 1000 images in six views that cover the 360$^\circ$ field of view (captured by the front, right-front, right-back, back, left-back and left-front cameras). 32-channel LiDAR data are also collected at 20 Hz. Human annotations are given in key LiDAR frames (one frame  is labeled in every 10 frames). 

We use both the unlabeled images and the LiDAR data for training our NeRF-LiDAR model. The labeled key LiDAR frames are used for evaluations.  
Limited by computing resources, 
we take 30 nuScenes sequences from the whole dataset. Each sequence covers a street scene with a length of  100$\sim$200m, and the total length is about 4km. Namely, we reconstruct about 4km of driving scenes for training and evaluation.

\paragraph{Evaluation Settings}
To avoid conflicts in the label definitions, we remap the image segmentation labels into five different categories (road, vehicles, terrain, vegetation and man-made) in accordance with the nuScenes LiDAR segmentation labels. 

In the training set, we use a total of 7000 unlabeled LiDAR frames and 30000 images for training our NeRF-LiDAR model. There are extra 1000 labeled LiDAR frames provided in these nuScenes scenes. We mainly use these labeled data for testing and fine-tuning in the experiments.

To evaluate the quality of the generated point clouds and point-wise labels, we train different LiDAR segmentation models (Cylinder3D \cite{cylinder3d} and RangeNet++ \cite{milioto2019rangenet++}) on the generated data and compare the segmentation model with those models trained on the real nuScenes LiDAR data(25k iterations). 

We use two evaluation sets to evaluate the 3D segmentation results. The first {\em Test Set 1} consists of 400 labeled real point clouds that is extracted from the 30 reconstructed scenes which are not used for training. This validation set is from the same scenes as the simulation data. 

The second {\em Test Set 2}  is the whole nuScenes validation set which consists of $\sim$5700 LiDAR point clouds from other nuScenes scenes (not including the 30 selected scenes). This is used to test the quality and generalization/transfer abilities of the simulation data. We test the trained model in other unseen scenes ({\em results are available in the supplementary}). 

\setlength{\tabcolsep}{2.5pt}
\renewcommand\arraystretch{1.2}
\begin{table}[H] 
\setlength{\abovecaptionskip}{-4pt}
\setlength{\belowcaptionskip}{-4pt}
%\vspace{-2mm}
\centering
\resizebox{1\linewidth}{!}{%
\large

\begin{tabular}{lcccccc}
\specialrule{.2em}{.1em}{.1em}
&\multicolumn{6}{c}{{\em Test Set 1} } \\

Training Set & road& vege. &terrain & vehicle &manmade &mIoU \\ 

\hline
% Real10k      &  &    & &  \\
CARLA &76.4&47.3& \XSolid&33.7&54.4& 52.9$^*$\\ \hline 
Real 1k& 96.2& 83.6& 83.1& 83.0& 86.4& 86.5    \\
Real 10k&97.0&83.6&84.5&89.3&\textbf{87.8}& 88.4    \\\hline
%Sim1k & & \\
Sim20k& 93.5 & 70.4& 77.6& 79.1&80.7 & 80.3  \\
%Sim10k + Real 100(other scenes)    &      &    \\
Sim20k + real1k &\textbf{97.1} & \textbf{84.1 }&\textbf{ 85.3} &  \textbf{92.2}&  86.9 & \textbf{89.1}  \\
\specialrule{.2em}{.1em}{.1em}
\end{tabular}
}
\vspace{2mm}
\caption{Evaluation an comparisons with the real LiDAR data and CARLA. $^*$ mean IoU of four classes.} 

\label{tab:samesceneevaluation}
 % \vspace{-3mm}
\end{table}

\subsection{Ablation Study}
To demonstrate the effectiveness of our method components, in Table \ref{tab:ablation}, we conduct experiments on different components of our NeRF-LiDAR.
We conduct ablation studies on 25 sequences of all data.

\setlength{\tabcolsep}{13pt}
\renewcommand\arraystretch{1}
\begin{table}[ht!] 

\setlength{\abovecaptionskip}{6pt}
\setlength{\belowcaptionskip}{-4pt}

\centering
\vspace{-2mm}
\resizebox{\linewidth}{!}{%

% \begin{tabular}{l|ccc|cc|c}

% \toprule
% & \multicolumn{3}{c|}{\textbf{Ray drop}} &\multicolumn{2}{c|}{\textbf{Feature loss}} & \multicolumn{1}{c}{\textbf{mIoU}}\\ 
%   &   no ray drop & random drop & learning drop & vgg loss & rangenet loss \\
% \midrule
% \multirow{6}{1.5cm}{Simulation}  &\Checkmark& & & &  &65.7\\
%  && \Checkmark& & & & 63.5\\
%  && &\Checkmark & &  &66.3\\

% \cmidrule{2-6}

%  && &\Checkmark & \Checkmark&  &66.5\\
% & & &\Checkmark & & \Checkmark &\bf{69.9}\\

% \bottomrule
% \end{tabular}
\begin{tabular}{ccc|cc|c}
\toprule
 \multicolumn{3}{c|}{\textbf{Raydrop}} &\multicolumn{2}{c|}{\textbf{Feature loss}} & \multicolumn{1}{c}{\textbf{mIoU}}\\ 
     no raydrop & random & learning & vgg  & rangenet  \\
\midrule
\Checkmark& & & &  &65.7\\
 & \Checkmark& & & & 63.5\\
 & &\Checkmark & &  &66.3\\

\cmidrule{1-6}

& &\Checkmark & \Checkmark&  &66.5\\
 & &\Checkmark & & \Checkmark &\bf{69.9}\\

\bottomrule
\end{tabular}
}
\vspace{0.5mm}
\caption{Ablations on different settings of ray drop and feature loss. Models are evaluated on validation set {\em Test Set 2}.} 
\vspace{-3mm}
\label{tab:ablation}
 
\end{table}

\paragraph{Effects of Raydrop}
\begin{figure*}[ht!]
\centering
\vspace{-2mm}
\begin{subfigure}[b]{0.17\linewidth}
%\begin{minipage}[t]{0.5\linewidth}
%\centering
\includegraphics[width=1\linewidth]{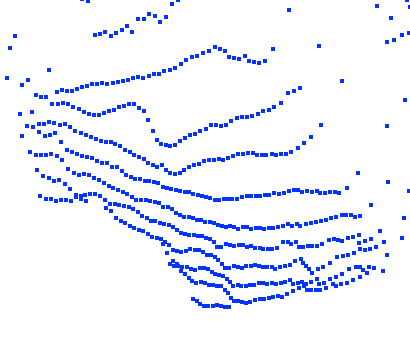}
\caption{Without RayDrop}

\end{subfigure}
\begin{subfigure}[b]{0.17\linewidth}
\centering
\includegraphics[width=1\linewidth]{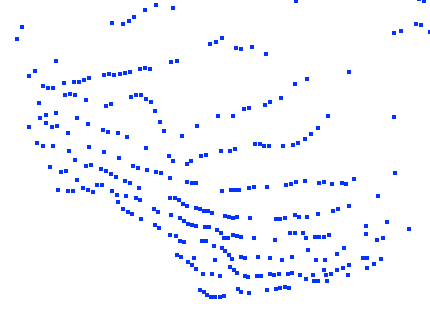}
%\caption{fig1}
\caption{Random RayDrop}
\end{subfigure}
\begin{subfigure}[b]{0.18\linewidth}
\centering
\includegraphics[width=1\linewidth]{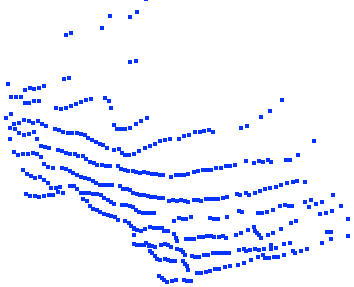}
\caption{Learning RayDrop}
%\label{subfig:our_nerf_lidar}
\end{subfigure}%
%\hspace{-2mm}
\begin{subfigure}[b]{0.19\linewidth}
\centering
\includegraphics[width=1\linewidth]{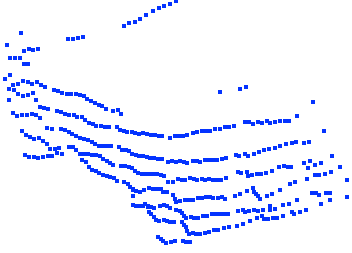}
\caption{W/ Feature Alignment}
\end{subfigure}%
\begin{subfigure}[b]{0.19\linewidth}
\centering
\includegraphics[width=1\linewidth]{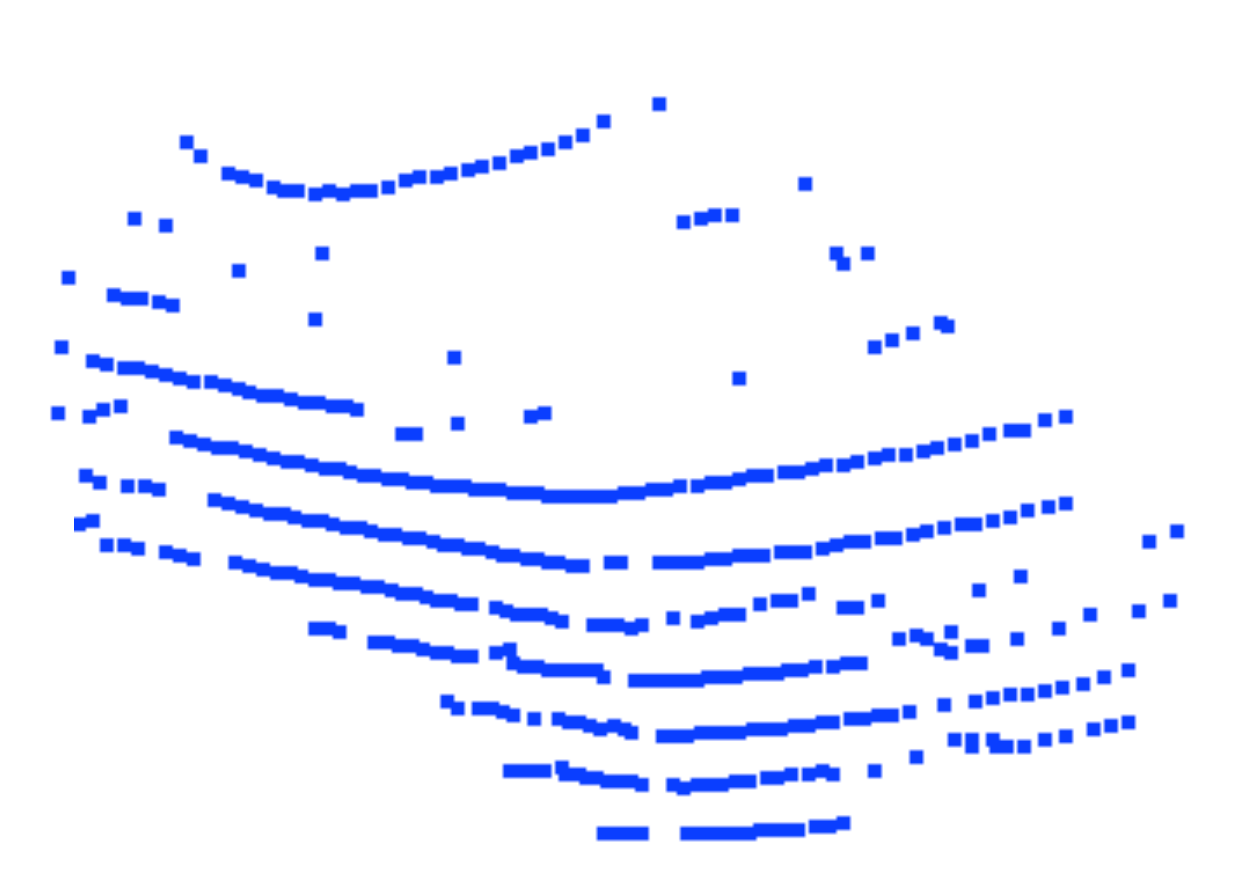}
\caption{Real LiDAR}
\end{subfigure}%

\caption{Comparisons of different settings for LiDAR rendering. (a) Point clouds without raydrop, (b) Point clouds after random raydrop. (c) Point clouds after our learning based raydrop but without using the feature-level alignment. (d) The final generated point clouds with both learning based raydrop and the feature-level alignment. 
(e) the real LiDAR point clouds.}
\label{fig:ablations}

\end{figure*}
We compare three different raydrop settings in Table \ref{tab:ablation} and Fig. \ref{fig:ablations}. Without any raydrop, the generated LiDAR data report a mean IoU of 65.7 after training the 3D segmentation model \cite{cylinder3d}. By adding a random drop strategy, the mean IoU is dropped to 63.5. And when we use our learning-based raydrop, the mean IoU is improved from 65.7 to 66.3.

\paragraph{Effects of Feature Loss}
We also evaluate the effects of the feature loss in Table \ref{tab:ablation}. We use two different feature extractors (VGG \cite{vgg} and RangeNet++ \cite{milioto2019rangenet++}) to implement the feature-level alignment. Without feature loss for feature-level alignment, our method reports a mean IoU of 66.3. By using the pre-trained VGG Net\cite{vgg} for feature alignment, the result is improved to 66.5. By implementing a pre-trained 3D segmentation network as the feature extractor, the results are significantly improved to 69.9.

\subsection{Label Quality}
\begin{figure*}[ht!]
\centering
\setlength{\abovecaptionskip}{4pt}
\setlength{\belowcaptionskip}{0pt}

\begin{subfigure}[b]{0.485\linewidth}
\begin{minipage}[t]{1\linewidth}
\centering
\includegraphics[width=1\linewidth]
{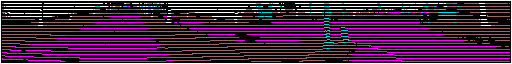}
\includegraphics[width=1\linewidth]
{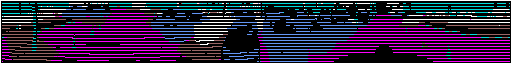}
%\caption{fig1}
\end{minipage}%
\caption{Our Data \& Label Rendering}
\end{subfigure}
\begin{subfigure}[b]{0.485\linewidth}
\begin{minipage}[t]{1\linewidth}
\centering
\includegraphics[width=1\linewidth]
{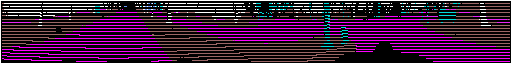}
\includegraphics[width=1\linewidth]
{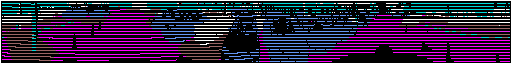}
%\caption{fig1}
\end{minipage}%
\caption{Ground-truth Labels of the Real Data}
\end{subfigure}
\caption{Comparisons between the data and label generated by the NeRF-LiDAR and the real LiDAR data with human annotations. For better visualization, we project the 3D point cloud as 2D equirectangular image with colorized labels.  Our NeRF-LiDAR (a) is shown able to generate accurate labels and realistic point clouds that is almost the same as the real data (b).}
\label{fig:label_quality}
\vspace{-2mm}
\end{figure*}
In Fig. \ref{fig:label_quality}, we visualize the generated LiDAR labels and compare them with the ground-truth labels in the real data. We can observe that the generated point clouds by our NeRF-LiDAR have strong point-to-point correspondence with the real data and labels. The label is accurate and close to the manual annotations. In the supplementary, we also evaluate the quality of the labels by computing the mIoU by comparing it with the manual labels. Our NeRF-LiDAR can generate accurate labels with a high mean IoU of 80$\sim$95 under different settings.  

\noindent\textbf{Pseudo Segmentation of Real LiDAR Scenes.} Compared with the pseudo segmentation labels on the real data, our NeRF-LiDAR can generate more rarely seen hard cases (in different view angles and routes) which have not been be collected by the dataset. These data significantly improve the diversity of the training dataset and boost the accuracy in training 3D segmentation models (Table~\ref{tab:lidar_pseudo_label}).

\setlength{\tabcolsep}{2.5pt}
\renewcommand\arraystretch{1.2}
\begin{table}[t]

\setlength{\abovecaptionskip}{1pt}
\setlength{\belowcaptionskip}{-10pt}
    \centering

\resizebox{\linewidth}{!}{
    \begin{tabular}{c|ccccc|c}
    \specialrule{.2em}{.1em}{.1em}
           & road & vege. & terrain & vehicle  & manmade & mIoU\\ \hline
LiDAR + pseudo Seg  &  70.6   &  39.4   &    30.7  &   20.9  & 52.0     &   42.7     \\ 

NeRF-LiDAR 10k&   \textbf{91.6}&  \textbf{61.1} &\textbf{ 59.2}& \textbf{ 69.7} & \textbf{68.0}&  \textbf{69.9}\\ 
\specialrule{.2em}{.1em}{.1em}

    \end{tabular}
    }
    \vspace{0.5mm}
    \caption{Comparisons between the NeRF-LiDAR data and the unlabeled real LiDAR data that uses pseudo segmentation labels.}
    \label{tab:lidar_pseudo_label}
    \vspace{-1mm}
\end{table}

\subsection{Comparisons with State of the Art} % Existing LiDAR Simulation Methods}
\noindent\textbf{CARLA and Real LiDAR}
In Table \ref{tab:samesceneevaluation}, we evaluate the quality of the generated data by using it to train 3D segmentation model \cite{cylinder3d}, the mean IoU is used as the evaluation metric. 

In Table \ref{tab:samesceneevaluation}, we use the predicted weak image segmentation labels and 1000 labeled LiDAR frames to train our NeRF-LiDAR. We use the {\em Test Set 1} which is extracted from the same scenes as the simulation data to evaluate the accuracy of the 3D segmentation models.  CARLA simulator \cite{carla} and different real LiDAR sets are taken as baselines. The real 1k and real 10k data contain the labeled frames in the scenes. When we train the point cloud segmentation network on the 20k simulation data, it achieves a mean IoU of 80.3, which is close to real 1k data and far better than the model trained on the 20k CARLA data. If we use 1k real data for fine-tuning, the mIoU can be further improved to 89.1, which exceeds real 10k data.

\noindent\textbf{Reconstruction-based Simulators}
Our NeRF-LiDAR possesses apparent advantages over other reconstruction-based LiDAR simulators \cite{lidarsim,fang2020augmented}. 
First, RGB images are used to assist the reconstruction in our NeRF-LiDAR, providing useful multi-view geometry information for more accurate reconstruction and label generation. Secondly, no manual annotations on the point clouds are required. We use NeRF representation to learn label generation. Multi-view images provide useful label information and geometry consistency to reduce the outlier labels.
Moreover, no mesh reconstruction of the real world is required.  NeRF-LiDAR can preserve more details of the real world by leveraging the RGB texture contents.
Finally, NeRF-LiDAR does not require dense point clouds for the reconstruction of the real world.  LiDARSim \cite{lidarsim} uses a 64-channel LiDAR and scans the street many times to achieve the dense point clouds for LiDAR simulation. Augmented LiDAR simulator \cite{fang2020augmented} utilizes a special and expensive LiDAR device to achieve the dense point clouds. As a comparison, NeRF-LiDAR relies more on the 2D images to achieve 3D geometry information.

\begin{figure*}[ht!]
\centering
\vspace{-10mm}
\begin{subfigure}[b]{0.3\linewidth}
%\begin{minipage}[t]{0.5\linewidth}
%\centering
\includegraphics[width=1\linewidth,trim={2cm 3cm 0 0},clip]{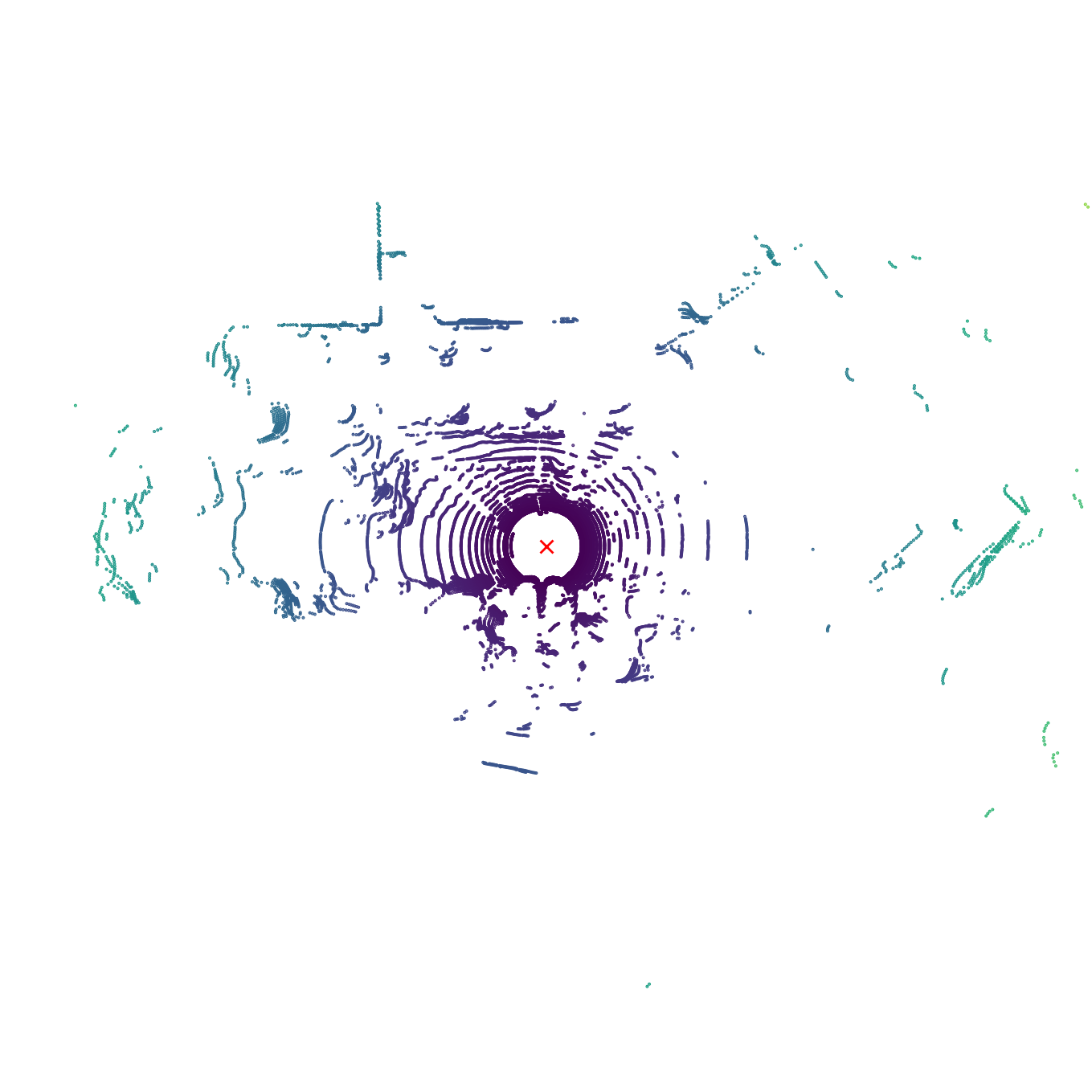}
\caption{LiDARSim \cite{lidarsim}}
% \label{subfig:carla}
\end{subfigure}
\hspace{2mm}
\begin{subfigure}[b]{0.3\linewidth}
\centering
\includegraphics[width=1\linewidth,trim={2cm 3cm 0 0},clip]{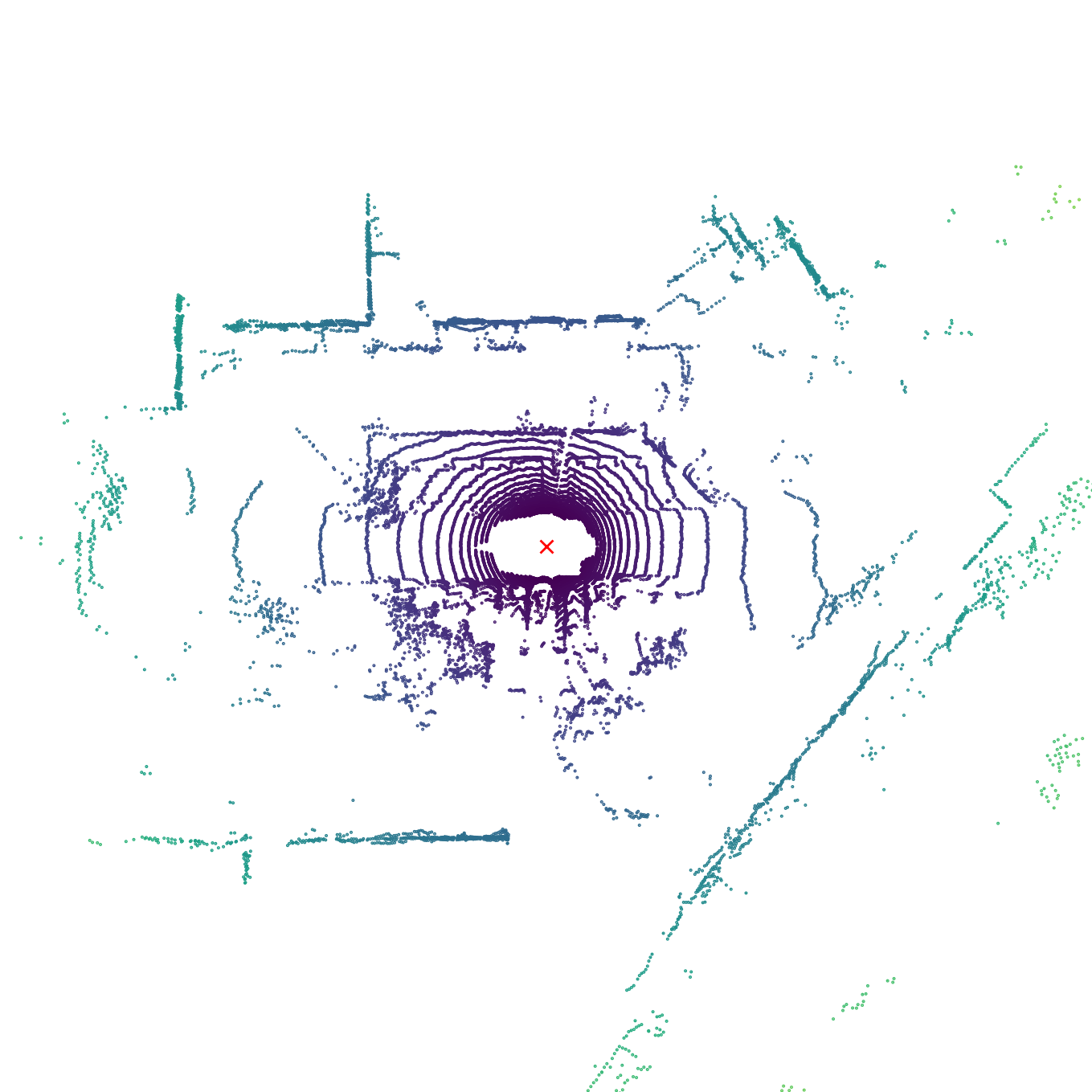}
\caption{Our NeRF-LiDAR}

\end{subfigure}
\hspace{2mm}
\begin{subfigure}[b]{0.3\linewidth}
\centering
\includegraphics[width=1\linewidth,trim={2cm 3cm 0 0},clip]{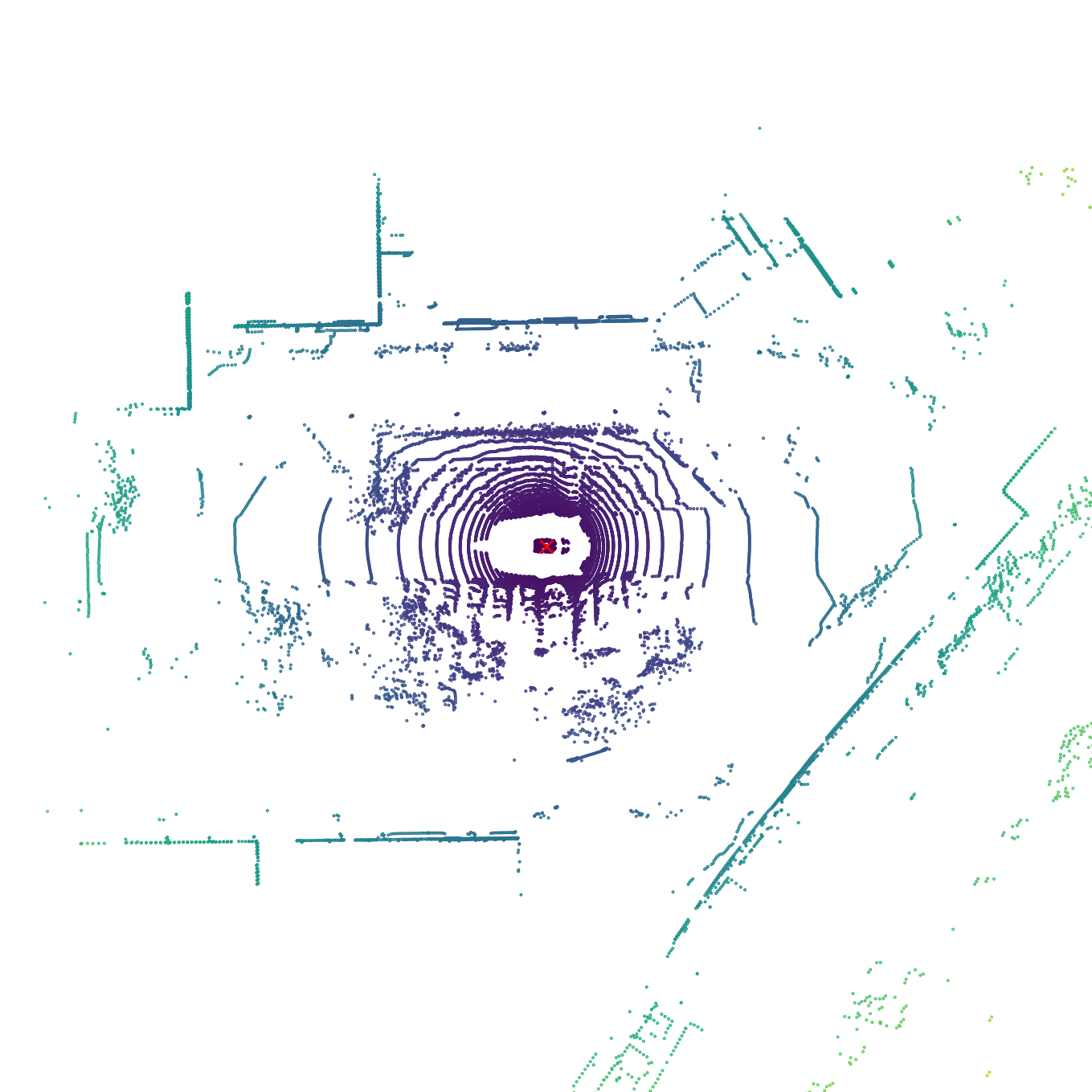}
\caption{Real LiDAR sensor}
\end{subfigure}

\caption{Comparison between our NeRF-LiDAR and LiDARSim~\cite{lidarsim}. Sparse aggregated point cloud leads to poor-quality mesh when reconstructing scenes. Thus, the simulated LiDAR points lose reality compared to NeRF-LiDAR.} 
 \label{fig:cmp_lidarsim}
\vspace{-2mm}
\end{figure*}

We compare our method with LiDARSim~\cite{lidarsim} in Table \ref{tab:compare_lidarsim} and Figure \ref{fig:cmp_lidarsim}. About 1km driving scenes are reconstructed for evaluations. Considering that the official code of LiDARSim is not published, we tried our best to reproduce the procedures, \ie accumulating the LiDAR points, calculating normals, building meshes and doing ray-casting and ray-dropping to generate simulation data. However, the aggregated points and labels on nuScenes dataset are not dense enough to build high-quality meshes and thus produce poor simulated results. 

\setlength{\tabcolsep}{2.5pt}
\renewcommand\arraystretch{1.2}
\begin{table}[t]
 \setlength{\abovecaptionskip}{5pt}
\renewcommand\arraystretch{1.2}

\setlength{\belowcaptionskip}{0pt}
\centering

\resizebox{1\linewidth}{!}{
    \begin{tabular}{c|ccccc|c}
    \specialrule{.2em}{.1em}{.1em}
           & road & vege. & terrain & vehicle  & manmade & mIoU\\ \hline
LiDARSim & 83.1 & 55.1 &39.1 &36.7 & 75.2&   57.8    \\ 
NeRF-LiDAR &  \textbf{92.5}&  \textbf{69.9} &\textbf{ 70.1}& \textbf{ 74.7} & \textbf{84.8}&  \textbf{78.4}\\ 
\specialrule{.2em}{.1em}{.1em}
    \end{tabular}
}
\vspace{0.5mm}
\caption{Comparison with LiDARSim. We reconstruct 25 sequences for both LiDARSim and our NeRF-LiDAR and generate 10k frames of LiDAR point clouds to train 3D segmentation models. The mean IoU is evaluated on the correspond part of these scenes in \em{Test set 1}}
\label{tab:compare_lidarsim}
\vspace{-6mm}
\end{table}
% \vspace{-3mm}

\noindent\textbf{Combining Real and Simulated Data}
We  combine real data with NeRF-LiDAR data generated from ground-truth scenes to see if simulated data can further boost performance when used for training. As shown in Table \ref{tab:samesceneevaluation}, simulated data along  with 10\% real data (real 1k) are able to achieve better performance (Sim10k + real 1k: 89.1) than 100\% real data (real 10k: 88.4).

\paragraph{Moving Objects and Object Placement}
We follow ~\cite{NSG} to model dynamic objects during the NeRF reconstruction stage to simulate the moving objects. We also reconstructed the cars according to the ~\cite{instant-neus} and then place them into the simulated scenes.

\iffalse
\subsection{Comparisons with Different NeRF Baselines}
There are also many other NeRFs that can be used for scene reconstruction. 
Most of them (\eg Neus~\cite{neus}, VolSDF~\cite{volsdf}) can only be used for object or small scene reconstruction. These methods fail in large-scale outdoor driving scenes. So we mainly make comparisons with the large-scale NeRFs (Urban-NeRF~\cite{urban-nerf} and Panoptic Neural Fields (PNF)~\cite{panoptic-neural}).\footnote{More detailed comparisons are available in the supplementary.} 

As shown in Table \ref{tab:efficiency}, our NeRF model is 20$\times$ faster than Urban-NeRF and PNF. Using Urban-NeRF or PNF to reconstruct 4km driving scenes would take about 60 days. That is a big limitation in real-world applications.
Our NeRF-LiDAR also produces more accurate depth rendering than PNF since PNF does not use point clouds in supervision.
\input{file/tables/compare_timing}
\fi

\paragraph{Efficiency}
We train a total of 100 NeRF-LiDAR blocks to represent the 4km 3D driving scenes. Each NeRF-LiDAR model is trained on 8 NVIDIA A40 GPU which takes about 2 hours. During rendering point clouds and labels, each LiDAR frame costs less than 1 second.

\section{Conclusion}
NeRF-LiDAR is proposed to generate realistic LIDAR point clouds  via learning neural implicit representation. Images are used in NeRF-LiDAR to compute accurate 3D geometry and labels rendering. The real LiDAR data is also utilized as supervision to learn more realistic point clouds. The effectiveness of our NeRF-LiDAR is verified by training 3D segmentation neural networks. It's shown that the 3D segmentation models trained on the generated LiDAR data can achieve similar mIoU as those models trained on the real LiDAR data.

\paragraph{Acknowledgments}
This work was supported in part by STI2030-Major Projects (Grant No. 2021ZD0200204), National Natural Science Foundation of China (Grant No. 62106050 and 62376060),
Natural Science Foundation of Shanghai (Grant No. 22ZR1407500) and 
USyd-Fudan BISA Flagship Research Program.

\bibliography{aaai24}
\newpage
\appendix 
\begin{center}
  \bfseries \LARGE APPENDIX 
\end{center}

\section{More Experimental Results and Details}
\subsection{LiDAR segmentation}
In our paper, we evaluate the quality of the generated LiDAR point clouds by training two popular 3D segmentation algorithms Cylinder3D \cite{cylinder3d} and RangeNet++ \cite{milioto2019rangenet++}. The evaluation results of Cylinder3D are reported in the paper (Table 1) and Table ~\ref{tab:fullvalidation}. In Table \ref{tab:rangenetvalidation}, we report the evaluation results of the RangeNet++ on different training sets.

We can observe that the LiDAR point clouds generated by our NeRF-LiDAR are  effective in training both the Cylinder3D and the RangeNet++. Especially, by using 20k simulation data for pre-training and another 1k real data for fine-tuning, the RangeNet++ achieves an improvement of 10\% in mean IoU when compared with the model trained on 1k real data and the model can achieve a better accuracy than 10k real data training. In Table ~\ref{tab:fullvalidation}, fine-tuning the pretrained model with 1k real data will boost the accuracy of foreground objects from 80.1 to 86.2.
This demonstrates the generalization of our NeRF-LiDAR and implies that our simulation LiDAR data can be used in many existing 3D segmentation networks and boost their training to achieve better accuracy. 
\setlength{\tabcolsep}{2.5pt}
\renewcommand\arraystretch{1.2}
\begin{table}[ht!] 
\setlength{\abovecaptionskip}{0pt}
\setlength{\belowcaptionskip}{-4pt}

\centering
\resizebox{1\linewidth}{!}{%
\large

\begin{tabular}{lcccccc}
\specialrule{.2em}{.1em}{.1em}
&\multicolumn{6}{c}{{\em Test Set 1} } \\

Training Set & ~road& vege. &terrain & vehicle &manmade &mIoU \\ 
\hline
% Real10k      &  &    & &  \\
CARLA&70.5&31.3&\XSolid&9.7&29.3& 35.2 $^*$   \\
\hline 

Real 1k&89.0&35.9&53.9&45.4&52.5&55.3      \\
Real 10k&89.9&40.6&59.0&\textbf{55.3}&56.2&60.2 \\ 
\hline
Sim20k&87.9&42.4&53.9&37.9&44.7&53.4   \\

Sim20k + real1k& \textbf{91.5} & \textbf{46.1} & \textbf{65.3} & 53.5 & \textbf{60.0} &  \textbf{63.3}  \\
\specialrule{.2em}{.1em}{.1em}
\end{tabular}
}
\vspace{4mm}
\caption{Evaluation on the {\em Test Set 1}, where the simulation data and the test set is from the same  driving scenes. Rangenet++\cite{milioto2019rangenet++} is trained on the different training sets for evaluations. $^*$ mean IoU of four classes.} 

\label{tab:rangenetvalidation}

\end{table}

\begin{table}[ht!] 
\setlength{\abovecaptionskip}{0pt}
\setlength{\belowcaptionskip}{-4pt}
%\vspace{-2mm}
\centering
\resizebox{1\linewidth}{!}{%
\large

\begin{tabular}{lcccccc}
\specialrule{.2em}{.1em}{.1em}
&\multicolumn{6}{c}{{\em Test Set 2} } \\

Training Set & ~road& vege. &terrain & vehicle &manmade &mIoU \\ 
\hline

Real 1k     &97.0&78.7&73.9&76.5&85.7& 82.4    \\
Real 10k&\textbf{97.8}&\textbf{81.3}&\textbf{82.1}&80.1&\textbf{87.6}&\textbf{85.8}    \\\hline

Sim20k + real1k&97.0&79.5&77.6&\textbf{86.2}&85.3&85.1   \\
\specialrule{.2em}{.1em}{.1em}
\end{tabular}

}
\vspace{4mm}

\caption{Evaluation on the {\em Test Set 2} (the nuScenes validation set), where the simulation data and the test set is from the different  driving scenes. $^*$ mean IoU of four classes.} 

\label{tab:fullvalidation}
 
\end{table}

\subsection{3D Detection Task}
We also conduct experiments on the 3D detection task. As shown in Table \ref{tab:detection}, the synthetic data achieved from our NeRF-LiDAR can significantly improve the 3D detection results. The average precision is improved by 7.3\% (from 41.0 to 44.0). This further demonstrates the effectiveness of our NeRF-LiDAR.

\iffalse
\setlength{\tabcolsep}{2.5pt}
\renewcommand\arraystretch{1.2}
\begin{table}[ht!]
\vspace{-1mm}
\setlength{\abovecaptionskip}{1pt}
\setlength{\belowcaptionskip}{-8pt}
\centering
\resizebox{1\linewidth}{!}{
    \begin{tabular}{c|cc}
    \specialrule{.2em}{.1em}{.1em}
           & IoU 0.5 & IoU 0.7 \\ \hline
simu &&\\
real 1k     &&   \\
real 10k &&   \\ 
 simu + real1k &&   \\ 
 \specialrule{.2em}{.1em}{.1em}
    \end{tabular}
    }
    \vspace{1mm}
    \caption{LiDAR Object Detection results}
    \label{tab:detection}
\end{table}
\fi 

\setlength{\tabcolsep}{20pt}

\begin{table}[t] 

\centering
\resizebox{0.95\linewidth}{!}{%
\large
\begin{tabular}{l|cc}
\toprule Training Data &  Average Precision (AP) \\ \hline
Real 1k & 41.0  \\
Simu 20k + real1k &  44.0 \\ 

\bottomrule
\end{tabular}
}
 \vspace{4mm}
\caption{Evaluation on the 3D detection task. We train a 3D vehicle detector CenterPoint\cite{yin2021center} on the real data as the baseline. Then, we pre-train the same detector model on the synthetic LiDAR data and further finetune the model on the real data. By introducing synthetic data in training the 3D detector. The average precision is significantly improved from 41.0 to 44.0. } 

\label{tab:detection}
 
\end{table}

\subsection{Performance Boost by Foreground-object Placement}
To boost the perception performance for the foreground objects, more 3D vehicles were placed into the background LiDAR scenes. Firstly, 3D meshes of the cars were reconstructed with the real-world images. It was followed by arrangement of cars on the road according to the rendered semantic information.  Finally point dropout are conducted to simulate the realistic cars in the LiDAR scenes.   As shown in Table \ref{tab:ablation_placement}, by placing more vehicles into the NeRF scenes, the vehicle detection results are significantly improved by 10.5\%.

\iffalse
\setlength{\tabcolsep}{2.5pt}
\renewcommand\arraystretch{1.2}
\begin{table}[ht!] 
\setlength{\abovecaptionskip}{0pt}
\setlength{\belowcaptionskip}{-4pt}
%\vspace{-2mm}
\centering
\resizebox{1\linewidth}{!}{%
\large
\begin{tabular}{l|cccccc}
\toprule Object placement & road  & vege. & terrain & vehicle & manmade & mIoU\\
\midrule
No placement& & &  & &  &\\
Ours &  &  &  &  &  & \\

\bottomrule
\end{tabular}
}
\vspace{4mm}
\caption{Effectiveness of object placement. We train the 3D vehicle detector CenterPoint\cite{yin2021center}   on the original NeRF-LiDAR data (9561 frames) and the augmented NeRF-LiDAR data with new cars inserted. The average precision is significantly improved by 10.5\%.} 
%\label{tab:ablation}
\label{tab:ablation_placement}
 
\end{table}
\fi
\setlength{\tabcolsep}{20pt}
\renewcommand\arraystretch{1.2}
\begin{table}[ht!] 
\setlength{\abovecaptionskip}{0pt}
\setlength{\belowcaptionskip}{-4pt}
%\vspace{-2mm}
\centering
\resizebox{1\linewidth}{!}{%
\large
\begin{tabular}{l|c}
\toprule Training  Setting &  Average Precision (AP)\\
\midrule
W/o object placement& 25.7 \\
W/ object placement &  28.4 \\

\bottomrule
\end{tabular}
}
\vspace{4mm}
\caption{Effectiveness of object placement. We train the 3D vehicle detector CenterPoint\cite{yin2021center}   on the original NeRF-LiDAR data (9561 frames) and the augmented NeRF-LiDAR data with 15654 new cars inserted. The average precision is significantly improved by 10.5\%.} 

\label{tab:ablation_placement}
 
\end{table}
\subsection{Optimization} 
 We optimize the NeRF-LiDAR model through the reconstruction loss $L_{rec}$ and the generation loss $L_{gen}$. 
 We set balance loss weights  as $w_{feat}=0.2,w_{rgb}=0.5$ and $\hat{w_l}=0.2$. 
 For a stable training and faster convergence, we first train the reconstruction for 40K iterations without using the generation loss. Then, the reconstruction part is frozen and we further learn the generation module for another 20k iterations. As for the reconstruction part, the batch size for sampling RGB rays is 2048 and the number  of the rays for LiDAR supervision is 2048. We adopt Adam optimizer to optimize our network. The learning rate is annealed log-linearly from  $5\times$ $10^{-4}$ to $5\times$ $10^{-6}$  with a warm-up phase of $2500$ iterations.
In the generation part, we adopts Adam with a learning rate of $1\times$ $10^{-4}$. The batch size of images is 4.
\subsection{Dynamics Objects Rendering}
To simulate the dynamic objecst in the stree scenes, we follow the method in Neural Scene Graph~\cite{NSG} to model instance objects. We adopt smaller model with fewer parameters to represent these foreground nodes. We use a hash table of resolution from $2^4$ to $2^{10}$ with one-level feature of grid for instance objects, while a hash table of resolution from $2^4$ to $2^{13}$ with four-level feature of grid for the background.
\subsection{More Ablation Studies}
\paragraph{Label quality of the generated point clouds}
In Table \ref{tab:ablation_label}, we test the label quality of our generated LiDAR data. 
In training our NeRF-LiDAR, both the weak image labels (computed by the pre-trained SegFormer \cite{segformer}) and the manually annotated semantic LiDAR point labels (if available) can be used in learning label generation. 
We design several experiments to test the label quality given different available labels for training.

In the first setting, only the weak image segmentation labels are used in training our NeRF-LiDAR for label rendering. We find that the weak image labels are effective in training our NeRF-LiDAR. Many label outliers can be resolved by constructing the multi-view geometry and consistency in the NeRF representation. It reports a mean IoU of 84.45 when compared with the ground-truth labels. This experiment implies that The generated labels are highly accurate and are effective in training 3D segmentation models (as demonstrated in Table 1 of the paper and Table 1\& 2 of this supplementary).  

In the second and the third settings, different amount of manual labels are provided. When using 20\% of the  manually labeled LiDAR point clouds (about 40 LiDAR frames) to boost the training of our NeRF-LiDAR, the label quality can be improved by 2.62\% in mean IoU. When more manual labels are provided for training (50\%, about 100 labeled LiDAR frames), the label accuracy can be further improved to 89.66\% (a 5.21\% improvement in mean IoU).

\setlength{\tabcolsep}{1.2pt}
\renewcommand\arraystretch{1.2}
\begin{table}[ht!] 
\setlength{\abovecaptionskip}{0pt}
\setlength{\belowcaptionskip}{-4pt}
%\vspace{-2mm}
\centering
\resizebox{1\linewidth}{!}{%
\large
\begin{tabular}{l|cccccc}
\toprule Label generation & road  & vege. & terrain & vehicle & m.m. & mIoU\\
\midrule
% raw label& 0.91174747& 0.81471594 &0.75776543& 0.86280414 &0.87524003 & 84.44546017008608\\
Image label& 91.17& 81.47 &75.78& 86.28 &87.52 & 84.45\\
% raw label + 20\% fine-tune & 0.93965239&  0.83215956&  0.8193958 &  0.87626315 & 0.88582199& 87.06585773430812 \\
Image label + 20\% LiDAR label & 93.96&  83.21&  81.94 &  87.63 & 88.58& 87.07 \\
% raw label + 50\% fine-tune &0.95806176 &0.85632262 &0.8750445&  0.89094615 &0.90252761& 89.65805289129533\\
Image label + 50\% LiDAR label &95.81 &85.63 &87.50& 89.09 &90.25& 89.66\\

% \midrule
% Ours
% &  & &   \\
\bottomrule
\end{tabular}
}
\vspace{4mm}
\caption{Ablations on label generation. There are 200 labeled LiDAR point clouds provided in the nuScenes dataset (from the 10 nuScenes scenes used in the experiments). Besides the weak image labels, we also use 0\%, 20\% and 50\% of these labeled LiDAR point clouds in training our NeRF-LiDAR for label generation.
Other labeled LiDAR point clouds (the rest $50$\%) is used to compute the mean IoU for evaluating the quality of the generated labels. For each point in the real point clouds, we assign it a label using our NeRF-LiDAR. This point label is compared with the ground-truth label to compute the mean IoU.} 
%\label{tab:ablation}
\label{tab:ablation_label}
 
\end{table}

\paragraph{Ablation on Balance Weights}
In the experiments, we set $w_{rgb} = 1.0, \hat{w}_{l} = 0.2$ and $w_{feat} = 0.2$ for the balance weights in Eq. (11) and Eq. (15). 
We further perform ablation studies on these balance weights in Table ~\ref{tab:ablation_balanced}.
For the RGB balance weights $w_{rgb}$, we find that $w_{rgb}=1.0$ is the best setting. Lower or higher values will lead to worse performance with a drop of $1.6\sim 3.6\%$ in mIoU.
For the semantic label weight $\hat{w}_l$ that is used 
for training label generation,  [0.2,1.0][0.2,1.0] is a reasonable range for the label balance weight $\hat{w}_l$.  To also preserve a high-quality RGB image rendering, we use default $\hat{w}_l=0.2$ (a smaller weight), instead of 1.0 (that can further improve the mean IoU by another $0.5\%$). 
And for the feature loss used in point cloud raydrop \& alignment, we use $w_{feat} =0.2$, which is the best setting. Using lower or higher values cannot achieve  better mean IoU in training 3D segmentation networks.

\setlength{\tabcolsep}{5pt}
\renewcommand\arraystretch{1.2}
\begin{table}[ht!] 
\setlength{\abovecaptionskip}{0pt}
\setlength{\belowcaptionskip}{-4pt}
%\vspace{-2mm}
\centering
\resizebox{1\linewidth}{!}{%
\large
\begin{tabular}{ccc|ccc|ccc|c}

\toprule  \multicolumn{9}{c|}{Balanced weights} & Segmentation results\\
\toprule
\multicolumn{6}{c|}{NeRF reconstruction} &\multicolumn{3}{c|}{Generation}&\\
%\cmidrule{1-9}

\multicolumn{3}{c|}{$w_{rgb}$}&\multicolumn{3}{c|}{$\hat{w}_l$}&\multicolumn{3}{c|}{$w_{feat}$}&mIoU\\
\cmidrule{1-9}
0.5 & 1 & 2& 0.05& 0.2 & 1  &0.05&0.2&0.5\\
\midrule
\checkmark&  &  & &\checkmark & & &\checkmark   & & 79.7 \\
& \underline{\checkmark}&  & & \underline{\checkmark}&  & & \underline{\checkmark} &  & \underline{81.5} \\
& & \checkmark & &\checkmark & & &\checkmark &    &77.7 \\
\midrule
& \checkmark&  & \checkmark& & &  &\checkmark  &  &81.3 \\
& \underline{\checkmark}&  & & \underline{\checkmark}&  & & \underline{\checkmark} &  & \underline{81.5} \\
& \checkmark&  & & &\checkmark   & & \checkmark &  &82.0  \\
\midrule
& \checkmark&  & & \checkmark&  &\checkmark &  &  & 80.1 \\
& \underline{\checkmark}&  & & \underline{\checkmark}&  & & \underline{\checkmark} &  & \underline{81.5} \\
& \checkmark&  & & \checkmark&  & && \checkmark   &  80.4\\
\bottomrule
\end{tabular}
}
\vspace{4mm}
\caption{Ablations on balanced weights.  We train the cylinder3d model \cite{cylinder3d} on the generated LiDAR point clouds. 2k simulation point clouds is generated. {\em Test set 1} is used for evaluation. The default setting is highlighted with underlines. } 
%\label{tab:ablation}
\label{tab:ablation_balanced}
 
\end{table}
\paragraph{LiDAR Depth Supervision}
In training our NeRF-LiDAR, unlabeled real LiDAR point clouds play an important role. LiDAR points provided extra 3D information to improve the accuracy of the NeRF reconstruction of the real world. Furthermore, it can supervise the NeRF-LiDAR model to make the generated point clouds more realistic.

In Table ~\ref{tab:ablation_lidar_depth}, we demonstrate the effectiveness of using unlabeled LiDAR point clouds to train our NeRF-LiDAR. We train our NeRF reconstruction with and without LiDAR depth supervision respectively. 
Then we generate simulation data and train 3D segmentation network \cite{cylinder3d}  on these generated data respectively. %The evaluation results are reported
When training without LiDAR depths (only images are used), the 3D segmentation network only achieves a mean IoU of 63.7\%. When the LiDAR depths are provided for training, the accuracy is significantly improved to 84.4\% (with an improvement of 20.7\% in mean IoU).

\begin{figure}[ht!]
    \centering
    \includegraphics[width = 1\linewidth]{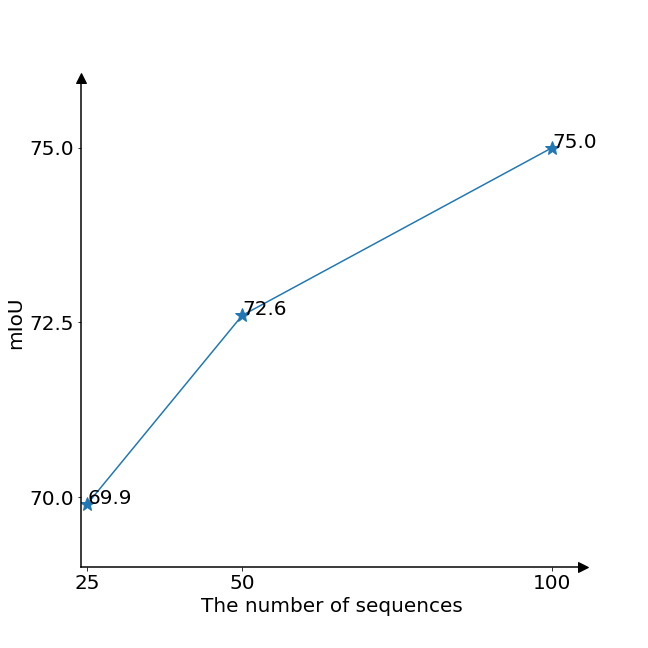}
    \caption{Ablation on the number of used sequences }
    \label{fig:ablation_scene_num}
\end{figure}

\setlength{\tabcolsep}{2.5pt}
\renewcommand\arraystretch{1.2}
\begin{table}[ht!] 
\setlength{\abovecaptionskip}{0pt}
\setlength{\belowcaptionskip}{-4pt}
%\vspace{-2mm}
\centering
\resizebox{1\linewidth}{!}{%
\large
\begin{tabular}{l|cccccc}
\toprule Depth supervision & road  & vege. & terrain & vehicle & manmade & mIoU\\
\midrule
No depth supervision&75.0 &54.4 &48.1  &71.4&69.5  &63.7\\
Ours & 89.3 & 80.0 & 74.5 & 91.1 & 87.2 & 84.4\\

\bottomrule
\end{tabular}
}
\vspace{4mm}
\caption{Ablations on the LiDAR depth supervision. LiDAR depth is used to reconstruct more accurate 3D geometry in learning the NeRF representation and also make the generated point clouds more realistic. We studied the effects of using  unlabeled real LiDAR point clouds in training our NeRF-LiDAR. } 
%\label{tab:ablation}
\label{tab:ablation_lidar_depth}
 
\end{table}
\paragraph{Ablation on the Number of Reconstructed Sequences}
Ablation studies was performed on the number of sequences used to verify the effectiveness of enriching the reconstructed scenes.  Referring to Fig. ~\ref{fig:ablation_scene_num} and Table ~\ref{tab:fullvalidation}, it revealed that reconstructing more scenes can improve the effect of simulation data. Only 30 scenes were reconstructed in the whole dataset due to the limited computing resource. Under the circumstances that nuScenes dataset contains over 1000 scenes, it was believed that the effect of simulation data would definitely be further improved with more scenes reconstructed.

\section{More Comparisons and Analysis}
\subsection{RGB Image and LiDAR simulation}
Our NeRF-LiDAR can simultaneously simulate the RGB novel views and the LiDAR point clouds. This makes it possible to develop and verify multi-sensor autonomous driving systems in our NeRF-LiDAR system.
Such multi-sensor simulation is impossible for any other existing LiDAR simulation methods. 

\begin{figure*}[ht]
\setlength{\abovecaptionskip}{-8pt}
\setlength{\belowcaptionskip}{10pt}
    \centering
\includegraphics[width=1.35\linewidth, trim={5cm 13cm 0cm 10cm},clip]{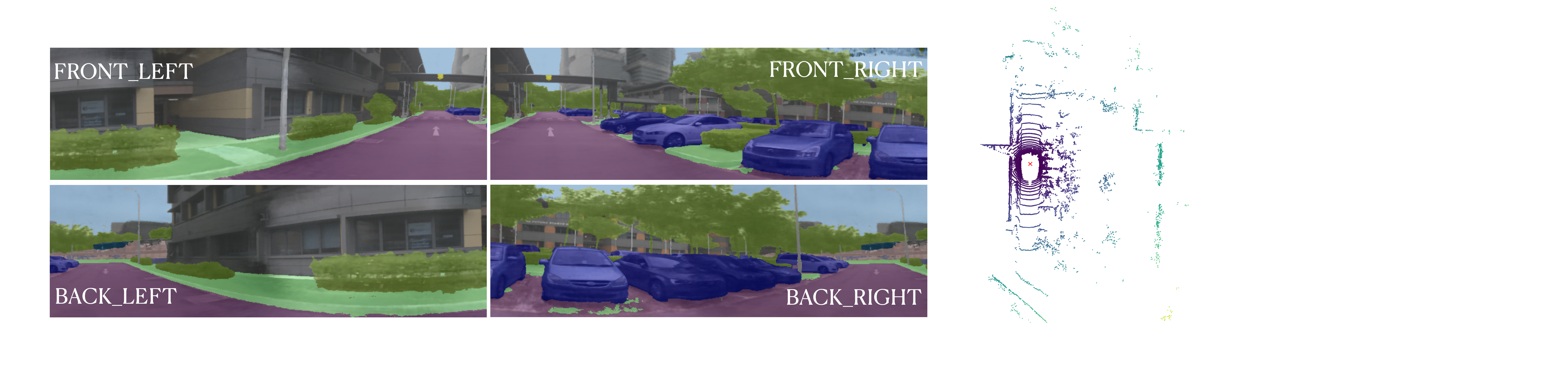}
    \caption{NeRF-LiDAR can also render high-quality RGB novel views along the the LiDAR. We simulate a 4 cameras and one LiDAR sensor configuration that is different from the neScenes configuration in the training dataset.  }
    \label{fig:rgb_lidar_rendering}
\end{figure*}

In Fig.  ~\ref{fig:rgb_lidar_rendering}, we present the simulation example of a novel sensor configuration which is different from the training data (nuScenes\cite{nuscenes2019}). This enables us to design and verify new sensor configurations and autonomous driving systems through the NeRF-LiDAR simulator.

\subsection{Comparison with other NeRF-based methods.} We implement Urban-NeRF and Panoptic Neural Fields on ten sequences and evaluate the results on the corresponding annotations. 
For Panoptic Neural Fields, since the source code is not available, we tried our best to reproduce the performance. 
As shown in Table~\ref{tab:ablation_nerf}, our method outperforms Urban-NeRF and the reproduced PNF on both the LiDAR segmentation task and the quality of point clouds.

\begin{table}[ht!] 
\setlength{\abovecaptionskip}{1pt}
\setlength{\belowcaptionskip}{-8pt}

\centering
\resizebox{0.8\linewidth}{!}{%
\large
\begin{tabular}{c|ccc}

   &  mIoU$\uparrow$ & MAE$\downarrow$& Recall@50$\uparrow$\\ \hline
Urban-NeRF & 56.9  & 0.862 & 0.815   \\
PNF & 59.9 & 0.968 & 0.763  \\
LiDARSim & 43.7 & 1.89 &  0.783 \\
Ours &  \textbf{66.7} & \textbf{0.483} & \textbf{0.892}\\
\end{tabular}
}
\caption{Comparison with other NeRF-based Methods. } 

\label{tab:ablation_nerf}
 
\end{table}

\subsection{Novel LiDAR View Rendering}

 New sensor settings can be synthesized for self-driving cars (\eg fewer or more video sensors), which is impossible for LiDARSim~\cite{lidarsim}.  
The sensors can be rotated and translated to generate rarely seen hard cases along with RGB rendering for video sensors.
\subsection{Visual Comparisons}

\begin{figure*}
\centering

% \begin{minipage}[t]{0.5\linewidth}
% \centering
% \includegraphics[width=1\linewidth]{figure/carla.png}
% \caption{Carla \cite{carla}}
% \end{minipage}%
\begin{subfigure}[b]{0.33\linewidth}
\begin{minipage}[t]{1\linewidth}
\centering
\includegraphics[width=1\linewidth]
{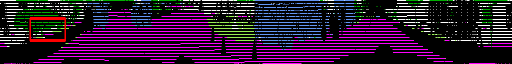}
\includegraphics[width=1\linewidth]
{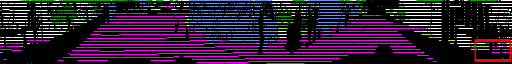}
\includegraphics[width=1\linewidth]
{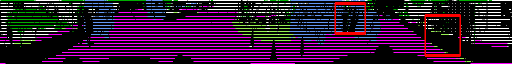}
\end{minipage}%
\caption{Real 1k}
\end{subfigure}
\begin{subfigure}[b]{0.33\linewidth}
\begin{minipage}[t]{1\linewidth}
\centering
\includegraphics[width=1\linewidth]
{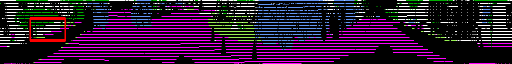}
\includegraphics[width=1\linewidth]
{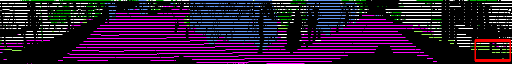}
\includegraphics[width=1\linewidth]
{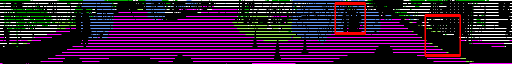}
%\caption{fig1}
\end{minipage}%
\caption{Sim 10k + real 1k}
\end{subfigure}
\begin{subfigure}[b]{0.33\linewidth}
\begin{minipage}[t]{1\linewidth}
\centering
\includegraphics[width=1\linewidth]
{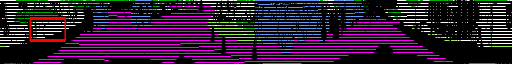}
\includegraphics[width=1\linewidth]
{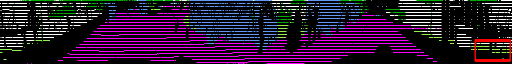}
\includegraphics[width=1\linewidth]
{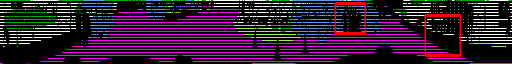}
%\caption{fig1}
\end{minipage}%
\caption{Ground truths}
\end{subfigure}
\caption{Visualization of the 3D segmentation results. Cylinder3D \cite{cylinder3d} is trained on different datasets for comparisons. Model (a) is trained on real 1k LiDAR data. Model (b) is pre-trained on the 10k simulation data generated by our NeRF-LiDAR and then fine-tuned on the 1k real data. (c) Ground truth lalels. As highlighted by the red windows, the LiDAR data generated by our NeRF-LiDAR can significantly improve the segmentation accuracy.  {\em Zoom in to see details} }
\label{fig:model_predicted_compares}
\end{figure*}
In Fig.  ~\ref{fig:model_predicted_compares}, we visualize the 3D segmentation results. Our simulated LiDAR point clouds can be used to boost the training of the 3D segmentation models. Compared with the model trained only with real data, the model that is pre-trained on our generated data and fine-tuned on the real data achieves far better segmentation accuracy. The segmentation accuracy of the five classes are all significantly improved.

\begin{figure*}[ht!]
\centering
%%%%%%%%%%%%%%%%%%% first fig
\begin{subfigure}[b]{0.22\linewidth}
\includegraphics[width=1\linewidth,angle=90]{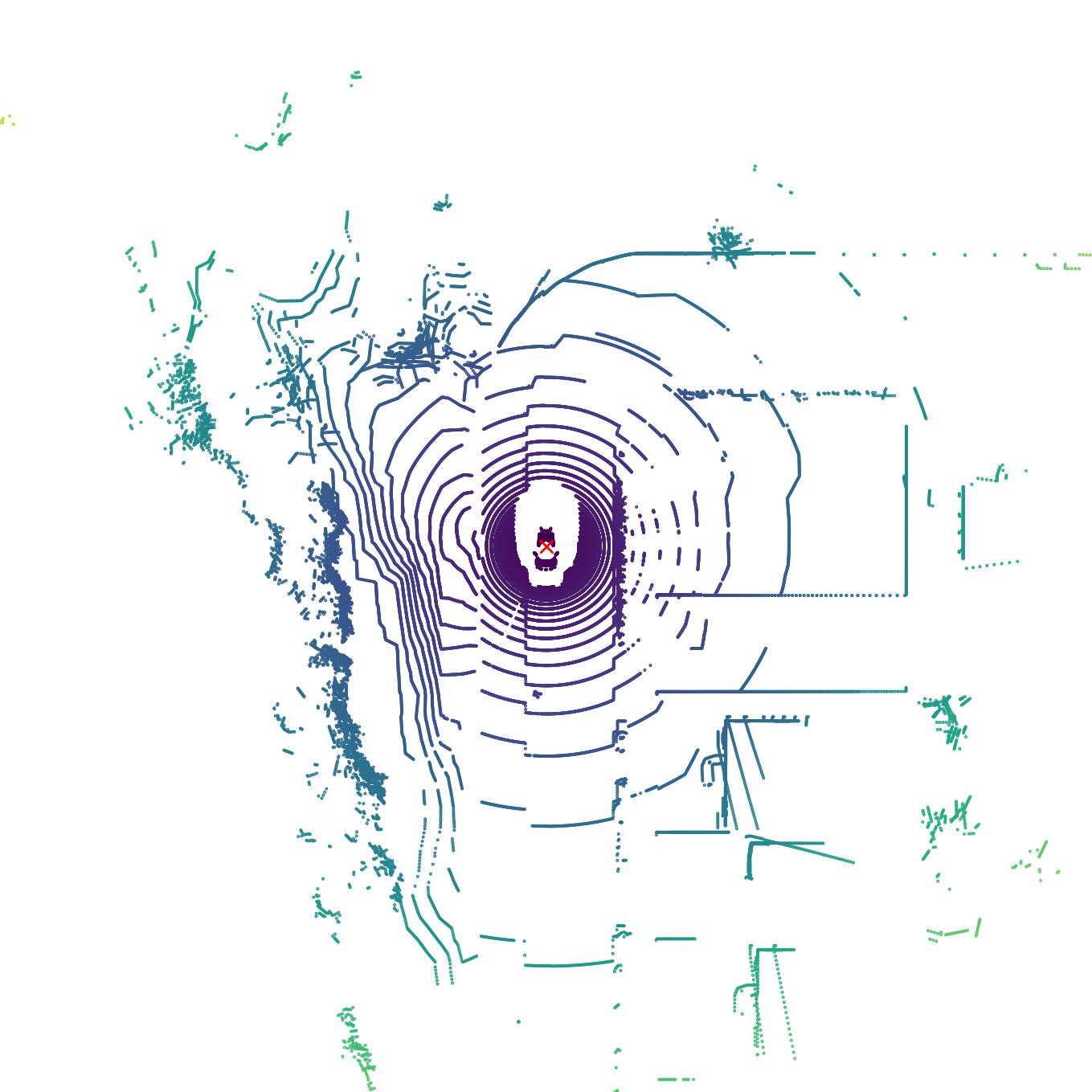}
\end{subfigure}
\begin{subfigure}[b]{0.22\linewidth}
\centering
\includegraphics[width=1\linewidth,angle=90]{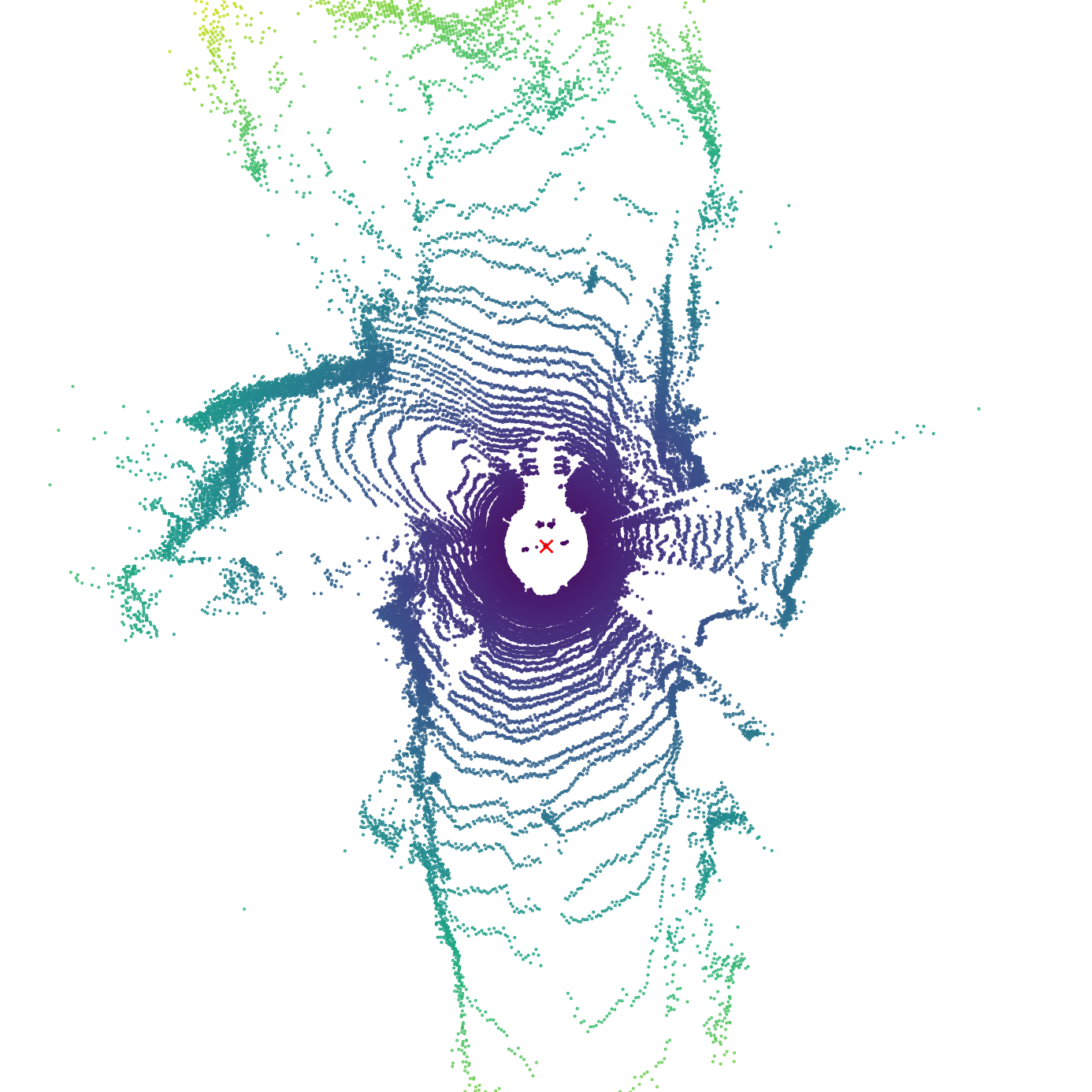}
\end{subfigure}
\begin{subfigure}[b]{0.22\linewidth}
\centering
\includegraphics[width=1\linewidth]{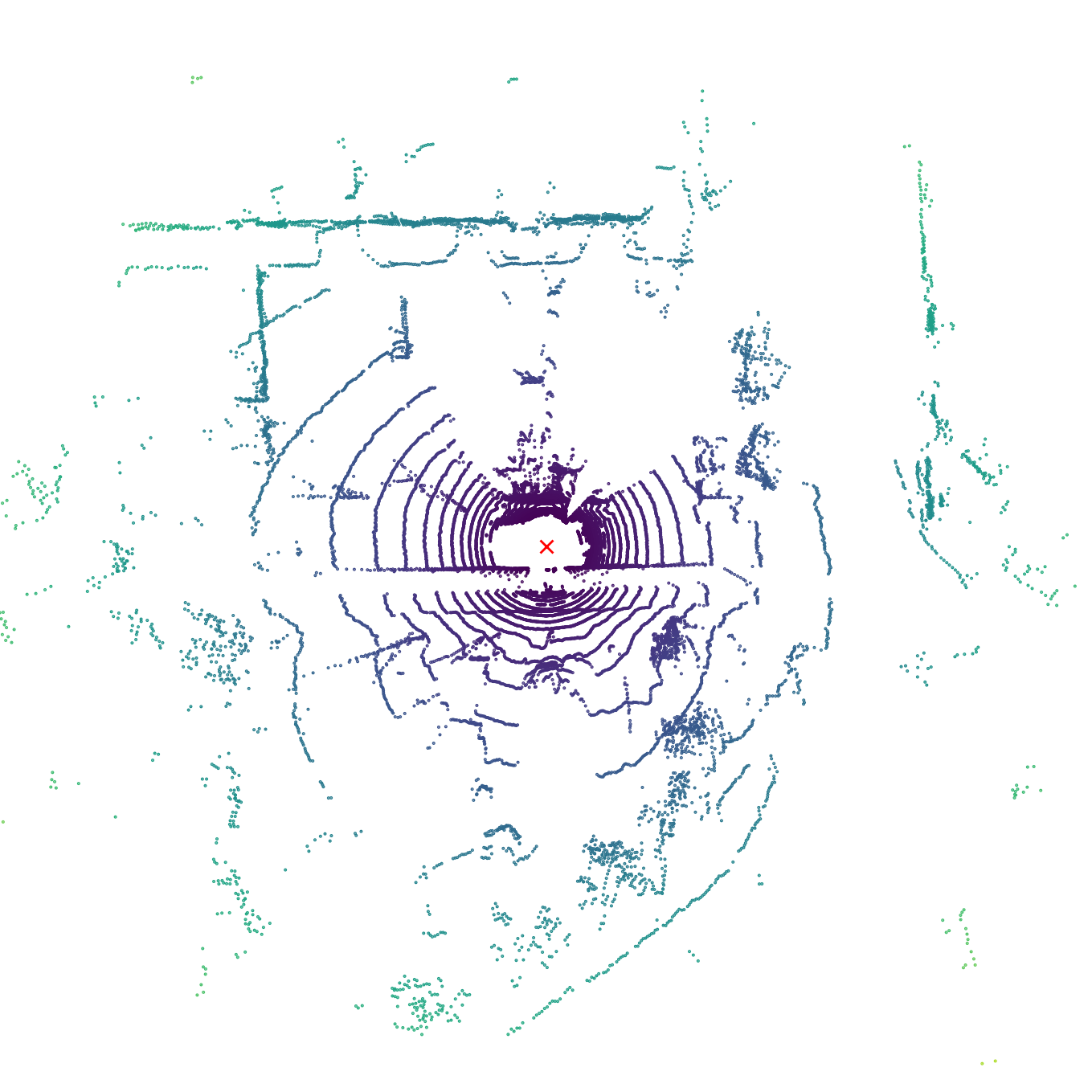}
\end{subfigure}
\begin{subfigure}[b]{0.22\linewidth}
\centering
\includegraphics[width=1\linewidth]{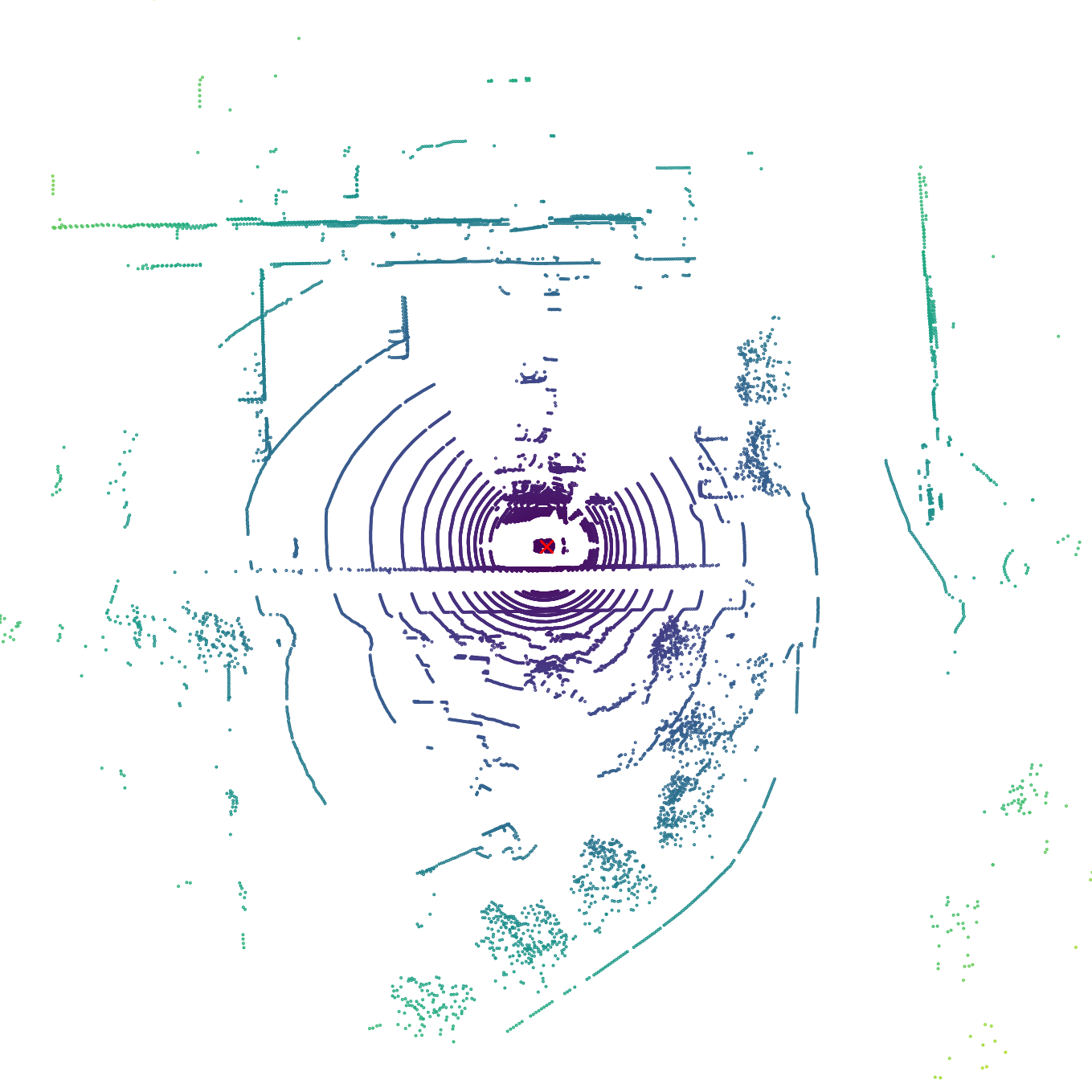}
\end{subfigure}\\
\begin{subfigure}[b]{0.22\linewidth}
\includegraphics[width=1\linewidth,angle=90]{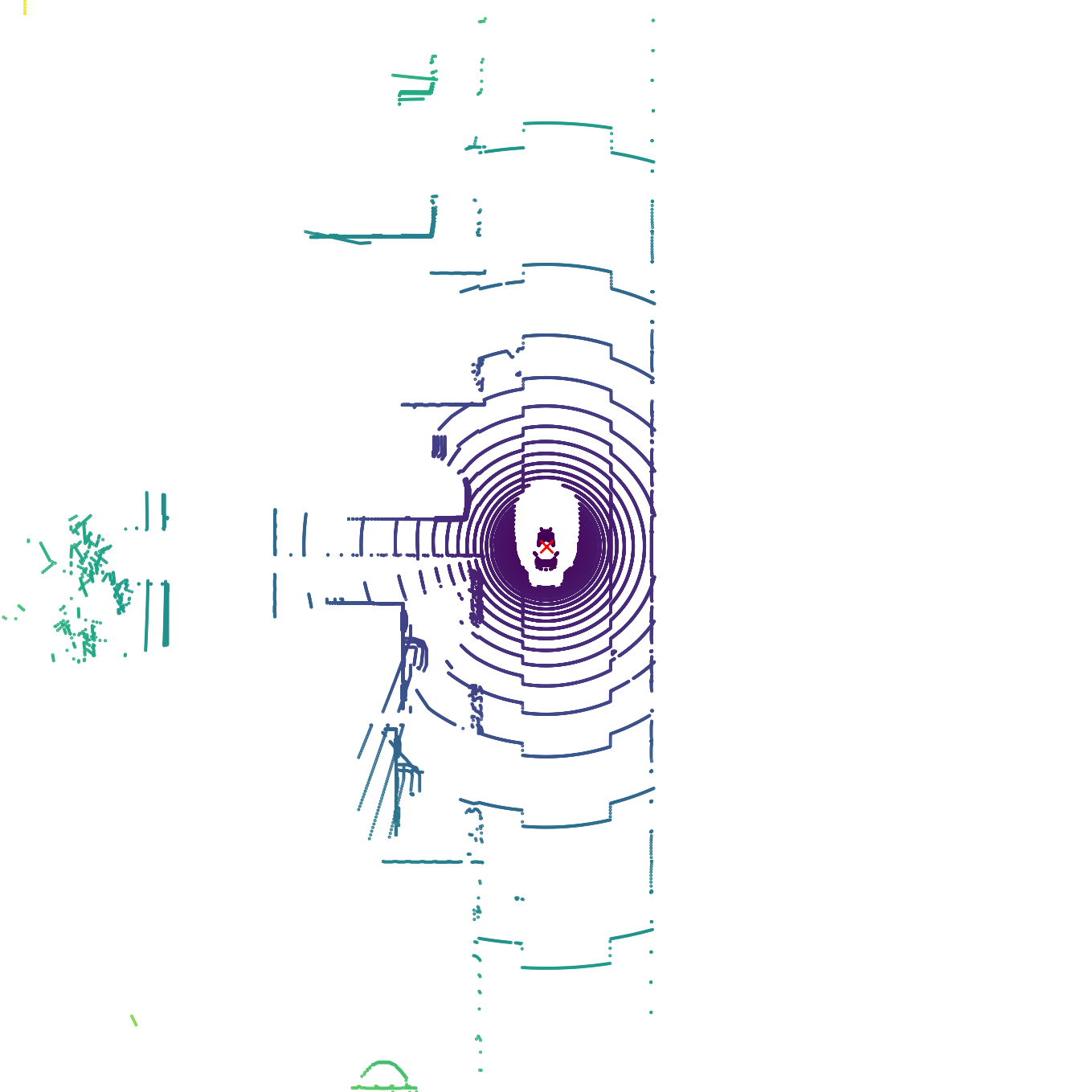}
\end{subfigure}
%%%%%%%%%%%%%%%%%%% second fig
\begin{subfigure}[b]{0.22\linewidth}
\centering
\includegraphics[width=1\linewidth,angle=90]{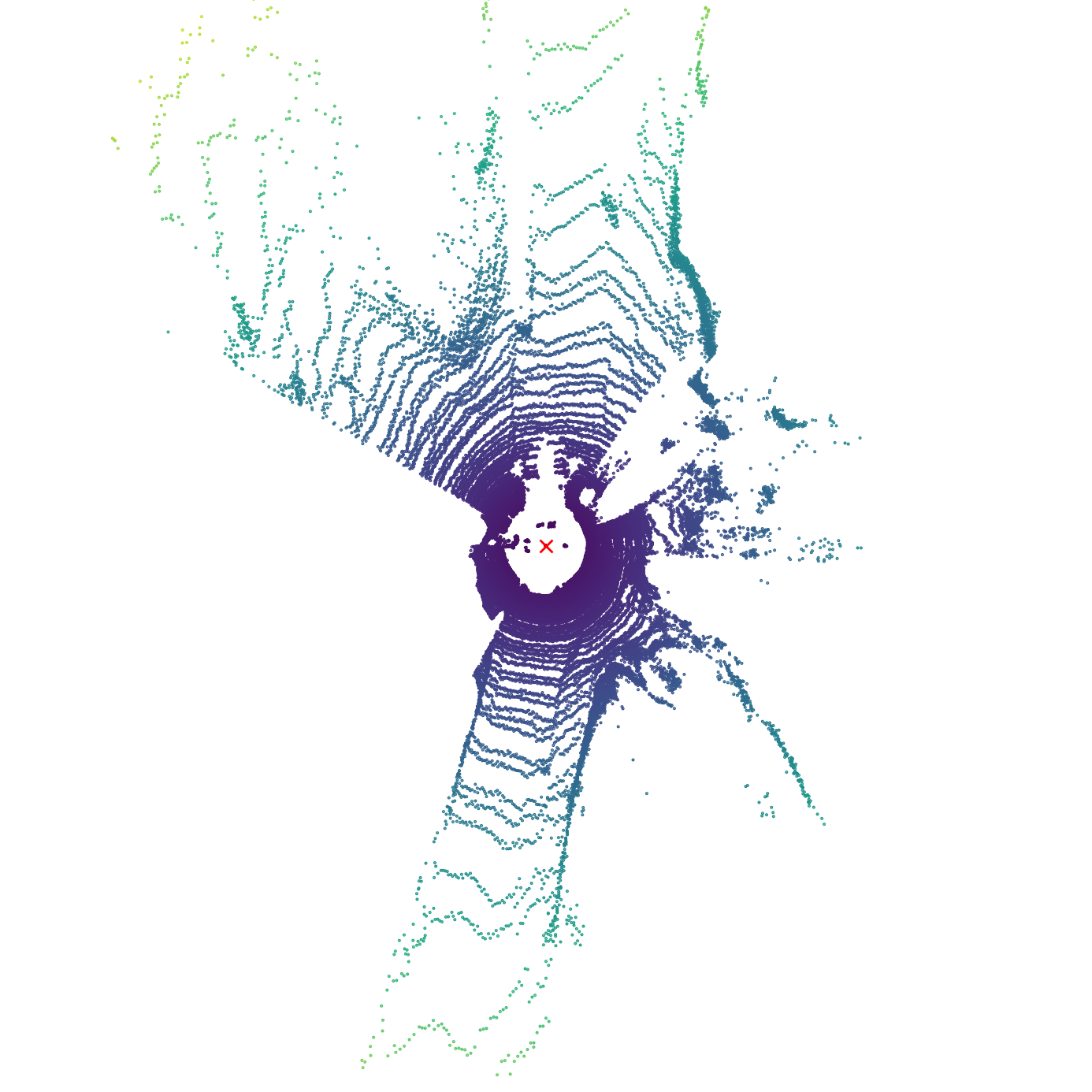}
\end{subfigure}
\begin{subfigure}[b]{0.22\linewidth}
\centering
\includegraphics[width=1\linewidth]{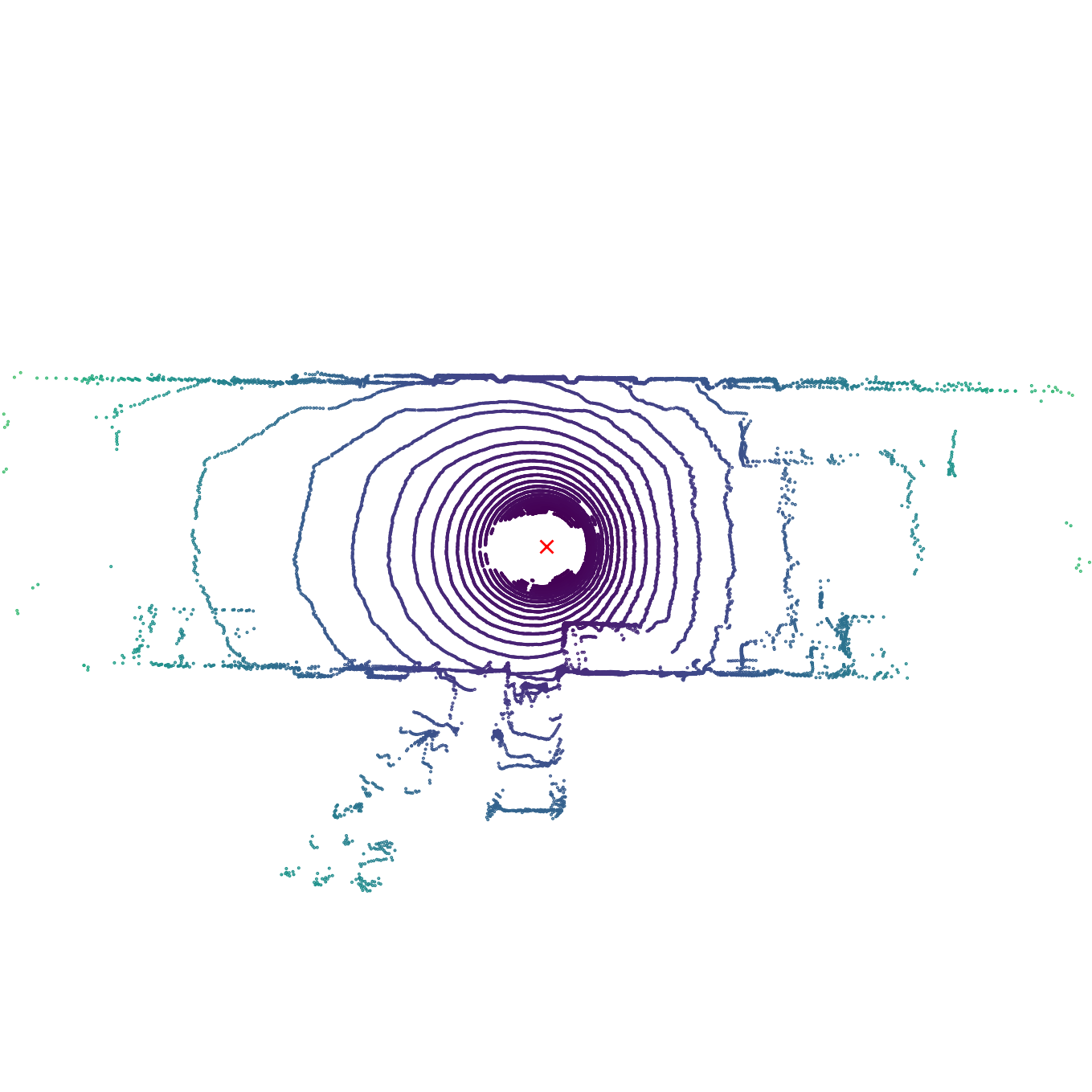}
\end{subfigure}
\begin{subfigure}[b]{0.22\linewidth}
\centering
\includegraphics[width=1\linewidth]{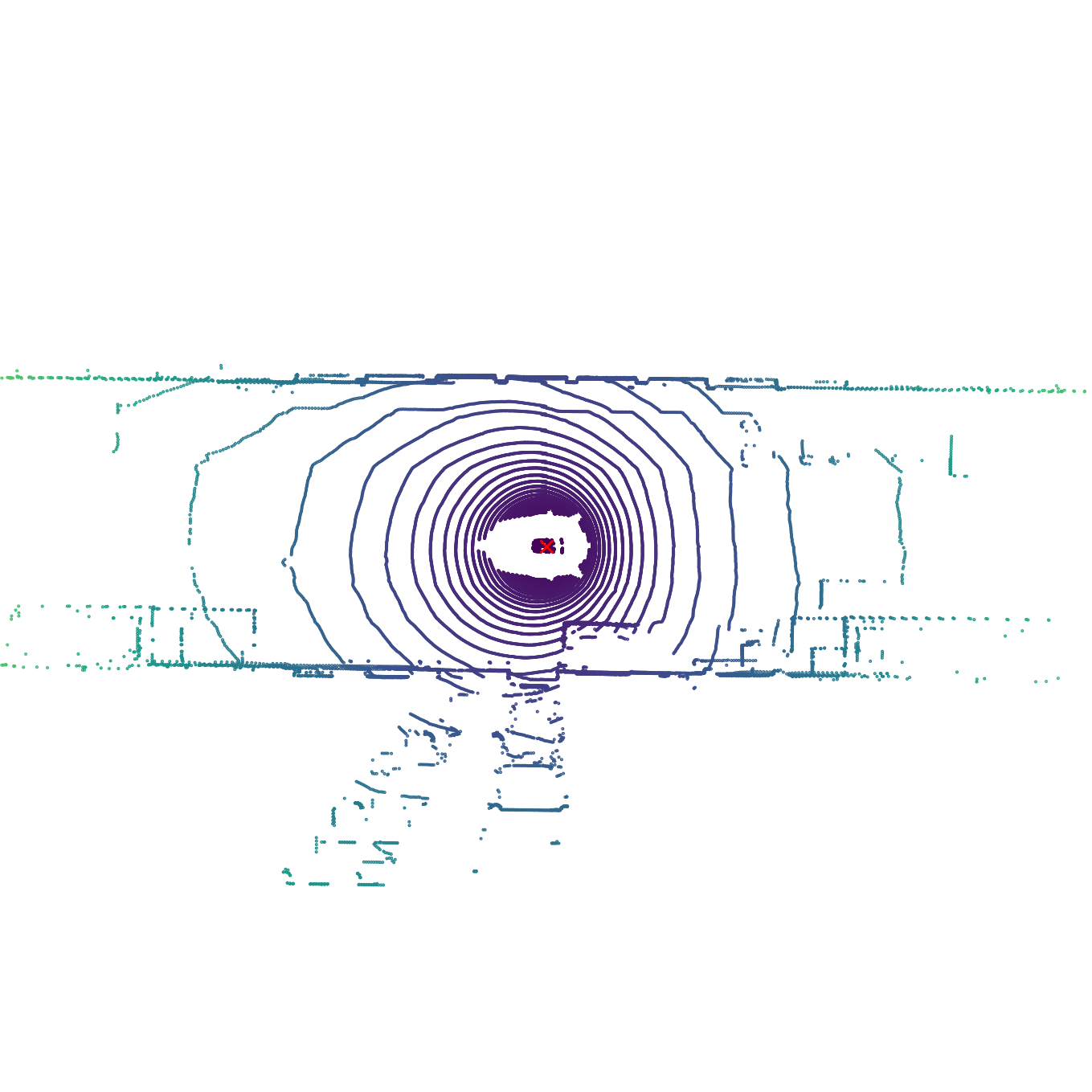}
\end{subfigure}\\
%%%%%%%%%%%%%%%%%%% third fig
\begin{subfigure}[b]{0.22\linewidth}
\includegraphics[width=1\linewidth,angle=90]{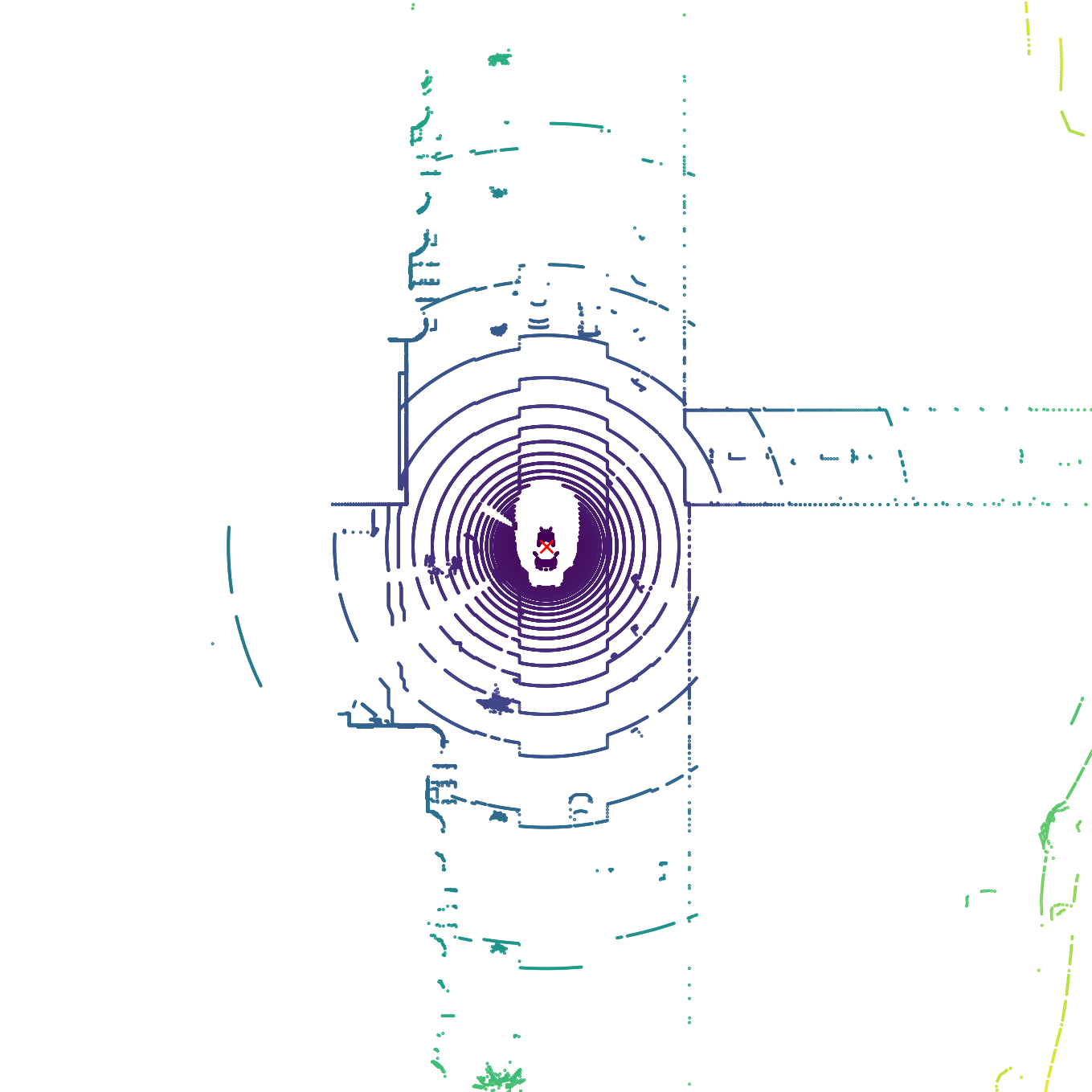}
\end{subfigure}
\begin{subfigure}[b]{0.22\linewidth}
\centering
\includegraphics[width=1\linewidth,angle=90]{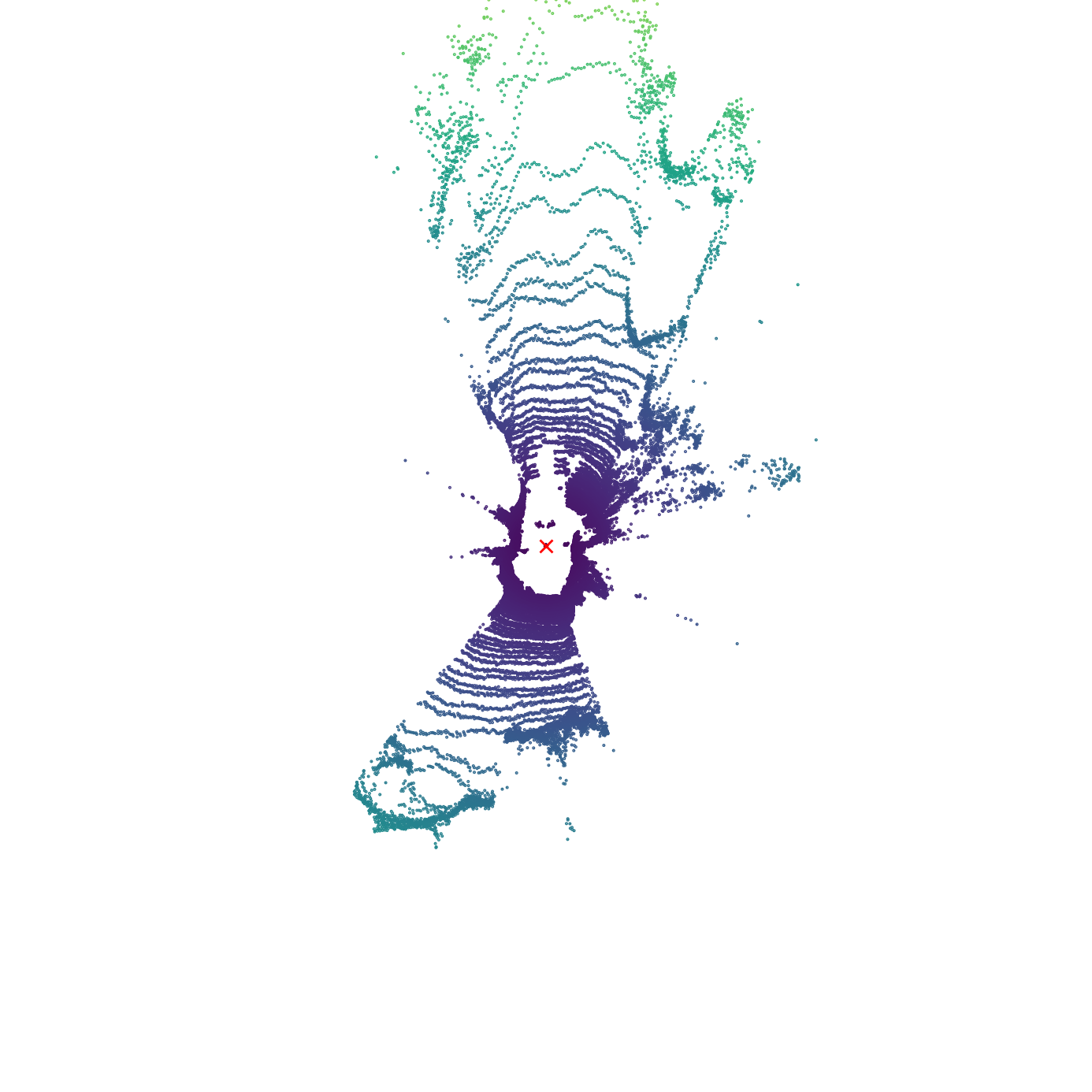}
\end{subfigure}
\begin{subfigure}[b]{0.22\linewidth}
\centering
\includegraphics[width=1\linewidth]{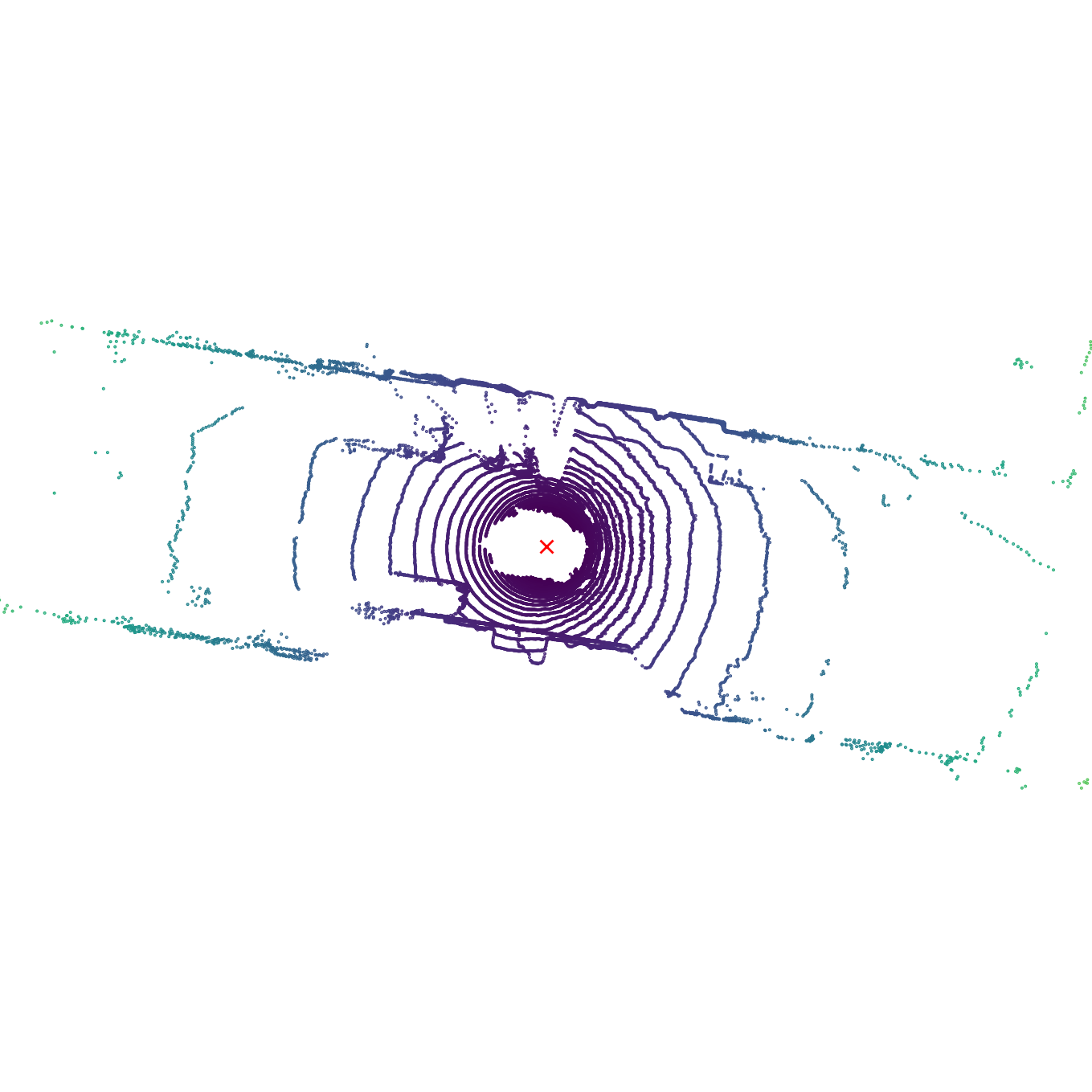}
\end{subfigure}
\begin{subfigure}[b]{0.22\linewidth}
\centering
\includegraphics[width=1\linewidth]{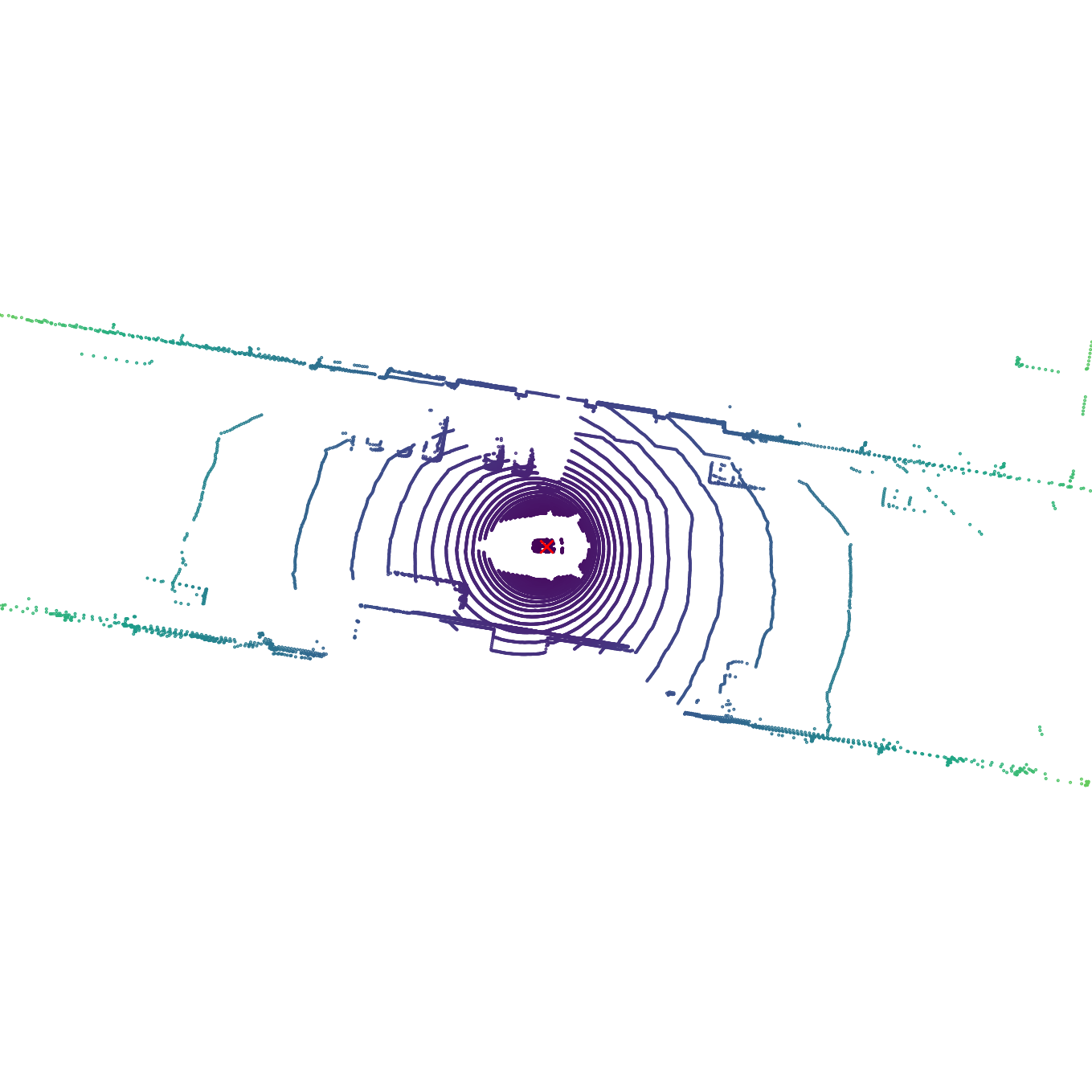}
\end{subfigure}\\
%%%%%%%%%%%%%%%%%%% fourth fig
\begin{subfigure}[b]{0.22\linewidth}
\includegraphics[width=1\linewidth,angle=90]{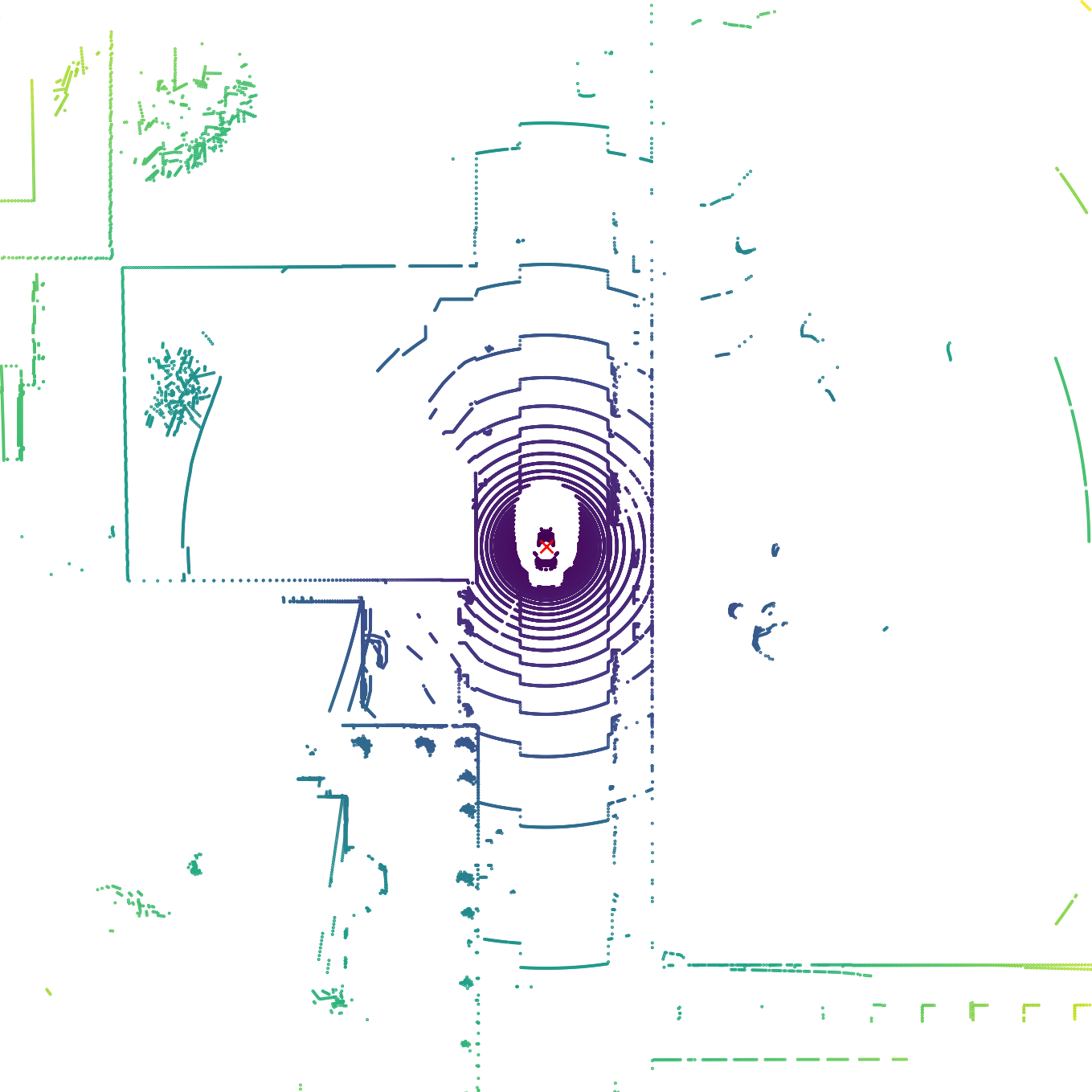}

\caption{CARLA \cite{carla}}
\label{sibfig:supp_lidar_carla}
\end{subfigure}
\begin{subfigure}[b]{0.22\linewidth}
\centering
\includegraphics[width=1\linewidth,angle=90]{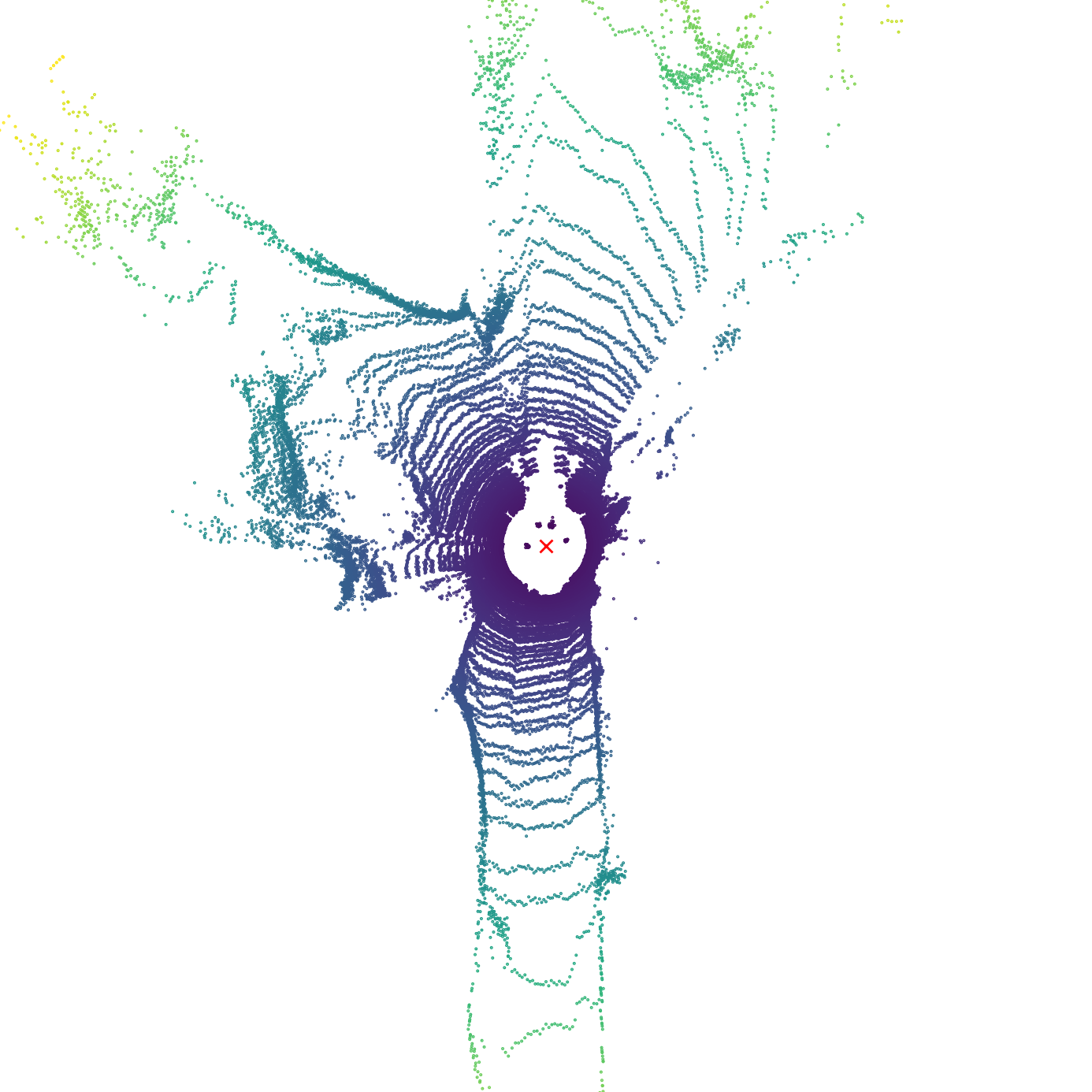}

\caption{LiDARGen~\cite{lidargen}}
\label{sibfig:supp_lidar_generative}
\end{subfigure}
\begin{subfigure}[b]{0.22\linewidth}
\centering
\includegraphics[width=1\linewidth]{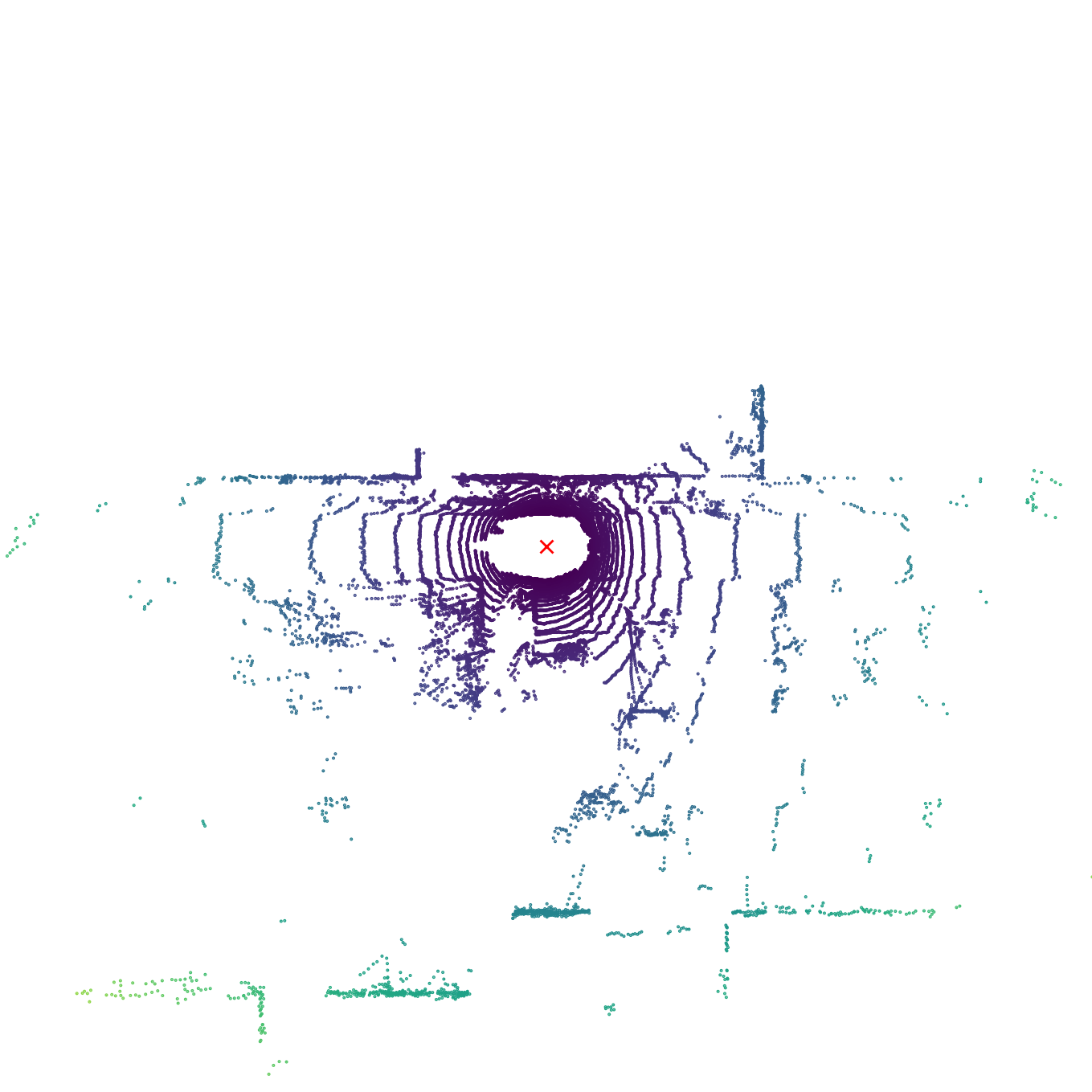}
\caption{Our NeRF-LiDAR}
\label{sibfig:supp_lidar_ours}
\end{subfigure}%
%\hspace{-2mm}
\begin{subfigure}[b]{0.22\linewidth}
\centering
\includegraphics[width=1\linewidth]{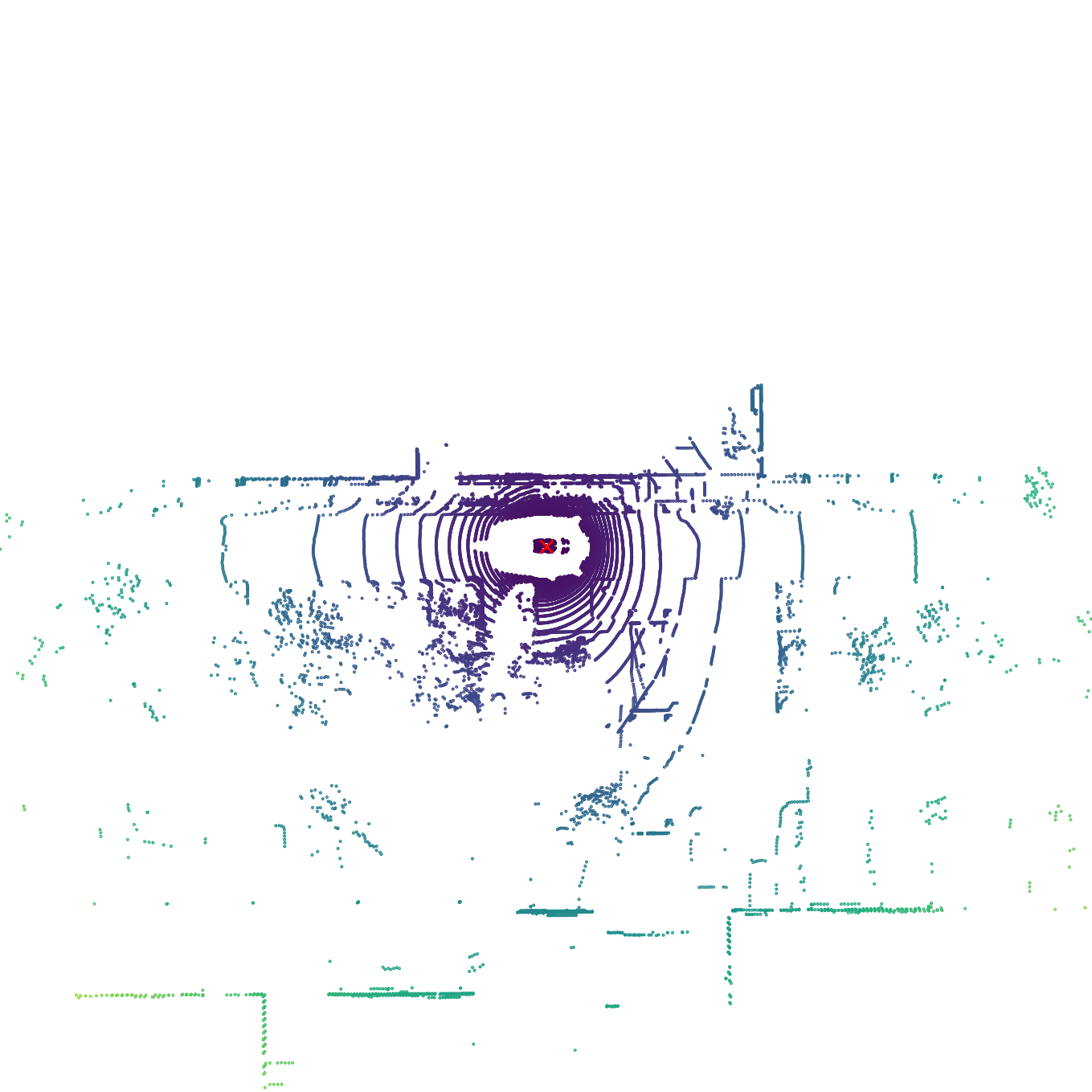}
\caption{Real LiDAR sensor}
\label{sibfig:supp_lidar_gt}
\end{subfigure}%\\
\caption{Comparisons of results between our NeRF-LiDAR and other existing LiDAR simulation methods. (a) Method \cite{carla} that creates virtual world for LiDAR simulation. (b) Diffusion model used for LiDAR generation \cite{lidargen}. (c) Our NeRF-LiDAR can generate realistic point clouds that is nearly the same as the real LiDAR point clouds (d). Real LiDAR sensor   {\em Zoom in to see details}}
\label{fig:performance_illustration_supp}
\end{figure*}

\subsection{Data Augmentation}
Apart from the pure simulation comparison (Table 4 of the main paper), we have additionally provided `LiDARSim + Real 1k' for consistent evaluation. 
\begin{table}[h!]
\vspace{-2mm}
    \centering
    \resizebox{1\linewidth}{!}{
    \begin{tabular}{c|cccccc}
     & road & vege.&  terrain & vehicle & manmade & mIoU \\ \hline
    Real 1k  &  96.2 & 83.6 & 83.1 & 83.0 & 86.4 & 86.5  \\
    LiDARSim + Real 1k  & 96.6 & 83.6 & 85.1 & 86.9 & 86.8 & 87.8 \\
    Ours + Real 1k & \textbf{97.1}& \textbf{ 84.1}&  \textbf{85.3 }& \textbf{92.2}&  \textbf{86.9 }& \textbf{89.1}

    \end{tabular}
    }
    \caption{Sim + Real experiments.}
    \vspace{-2mm}
    \label{tab:sim_real}
\end{table}
\subsection{NeRF-LiDAR vs Virtual-world Creation}
Compared with the LiDAR simulators (\eg CARLA \cite{carla}, AIRSIM \cite{shah2018airsim}) that is based on virtual-world creation, our NeRF-LiDAR leverages the real-world information for point-cloud simulation. As shown in Fig. ~\ref{fig:performance_illustration_supp}, Our LiDAR-NeRF can generate point clouds that are almost the same as the real LiDAR data (Fig. ~\ref{sibfig:supp_lidar_ours}). 
 CARLA (Fig. ~\ref{sibfig:supp_lidar_carla}) tends to generates evenly distributed LiDAR points. This is because the manually created 3D virtual world cannot simulate the complexity of the real world. 
Guillard \etal \cite{guillard2022learning} learns a deep neural network for raydrop to make the CARLA LiDAR simulator more realistic. However it still cannot simulate the complex details of the real world, such as the scene organization, object distributions and various shapes, surfaces and sizes of the real world contents.

\subsection{NeRF-LiDAR vs Generative Models}
The generative models utilize generative adversarial networks (GAN) \cite{li2018point,sauer2021projected,caccia2019deep} or diffusion model \cite{lidargen} for point cloud generation. It utilize a large number of real LiDAR point clouds to train the LiDAR generators. Even though these generated point clouds are visually similar to the real point clouds, they  cannot be used to train deep neural networks for segmentation or detection. This is because they have significant domain gaps with the real LiDAR point clouds in details (as compared in Fig. ~\ref{sibfig:supp_lidar_generative}) and the existing LiDAR generative models cannot generate accurate point labels for training 3D segmentation models. 

\end{document}